\documentclass{article}

\usepackage{microtype}
\usepackage{graphicx}
\usepackage{multirow}
\usepackage{booktabs} 
\usepackage{stfloats}
\usepackage{graphicx}
\usepackage{float}
\usepackage{subfig}
\usepackage{overpic}
\usepackage{hyperref}

\usepackage[accepted]{icml2024}

\usepackage{amsmath}
\usepackage{amssymb}
\usepackage{mathtools}
\usepackage{amsthm}

\usepackage[capitalize,noabbrev]{cleveref}
\theoremstyle{plain}
\newtheorem{theorem}{Theorem}[section]

\theoremstyle{definition}
\newtheorem{definition}[theorem]{Definition}

\theoremstyle{remark}

\usepackage[textsize=tiny]{todonotes}

\icmltitlerunning{Bias Mitigating Few-Shot Class-Incremental Learning}

\begin{document}

\twocolumn[
\icmltitle{Bias Mitigating Few-Shot Class-Incremental Learning}




\begin{icmlauthorlist}
\icmlauthor{Li-Jun Zhao}{sch}
\icmlauthor{Zhen-Duo Chen}{sch}
\icmlauthor{Zi-Chao Zhang}{sch}
\icmlauthor{Xin Luo}{sch}
\icmlauthor{Xin-Shun Xu}{sch}
\end{icmlauthorlist}

\icmlaffiliation{sch}{School of software, Shandong
University, Jinan, China}

\icmlcorrespondingauthor{Zhen-Duo Chen}{chenzd.sdu@gmail.com}

\icmlkeywords{}

\vskip 0.3in
]



\printAffiliationsAndNotice{}  

\begin{abstract}
Few-shot class-incremental learning (FSCIL) aims at recognizing novel classes continually with limited novel class samples. A mainstream baseline for FSCIL is first to train the whole model in the base session, then freeze the feature extractor in the incremental sessions. Despite achieving high overall accuracy, most methods exhibit notably low accuracy for incremental classes. Some recent methods somewhat alleviate the accuracy imbalance between base and incremental classes by fine-tuning the feature extractor in the incremental sessions, but they further cause the accuracy imbalance between past and current incremental classes. In this paper, we study the causes of such classification accuracy imbalance for FSCIL, and abstract them into a unified model bias problem. Based on the analyses, we propose a novel method to mitigate model bias of the FSCIL problem during training and inference processes, which includes mapping ability stimulation, separately dual-feature classification, and self-optimizing classifiers. Extensive experiments on three widely-used FSCIL benchmark datasets show that our method significantly mitigates the model bias problem and achieves state-of-the-art performance.
\end{abstract}

\section{Introduction}
\label{sec:intro}

In the dynamic and open real world,
Class-Incremental Learning (CIL)\cite{DBLP:conf/cvpr/RebuffiKSL17, DBLP:conf/icml/WenCQW0023} is proposed to continuously learn new emerging concepts and not forget the learned ones.
However, humans can establish new concepts
with only a few new examples when they have a certain amount of knowledge.
Therefore, Few-Shot Class-Incremental Learning (FSCIL)\cite{DBLP:conf/cvpr/GidarisK18, DBLP:conf/icml/AchituveNYCF21} is proposed to continuously learn novel classes with limited novel class samples after training on base classes with sufficient samples.

\begin{figure}[t]
  \centering
     \subfloat[CE loss (Base)]
     {\includegraphics[width=0.4\linewidth]{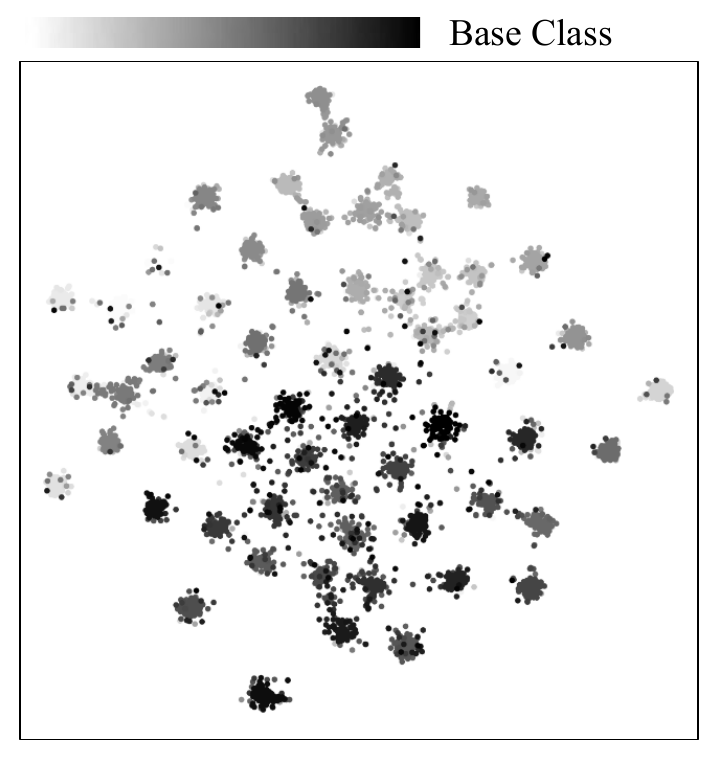}
    \label{fig:overlapa}} 
    \subfloat[CE loss (Base+Inc.)]{\includegraphics[width=0.4\linewidth]{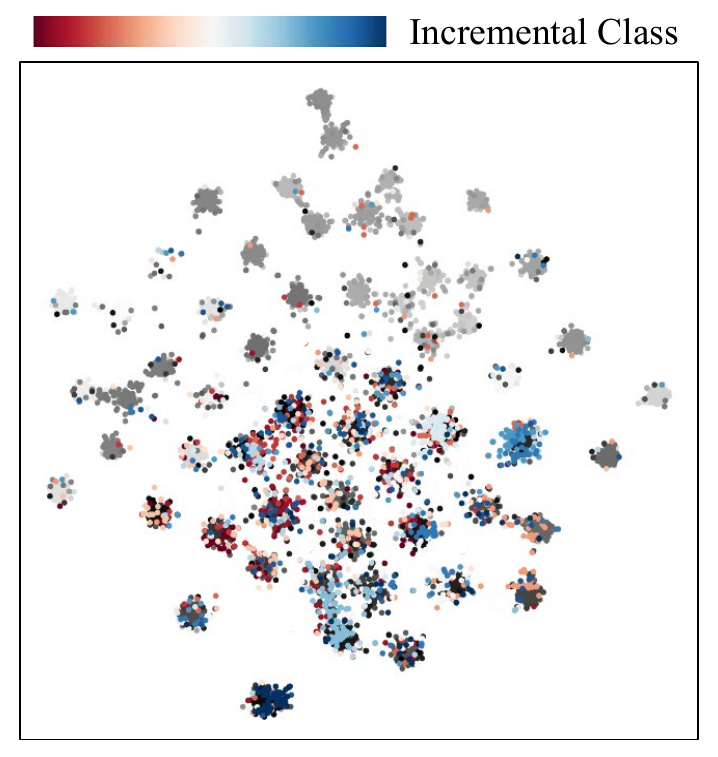}
    \label{fig:overlapb}}

    \subfloat[CE loss+$\delta $ (Base)]{\includegraphics[width=0.4\linewidth]{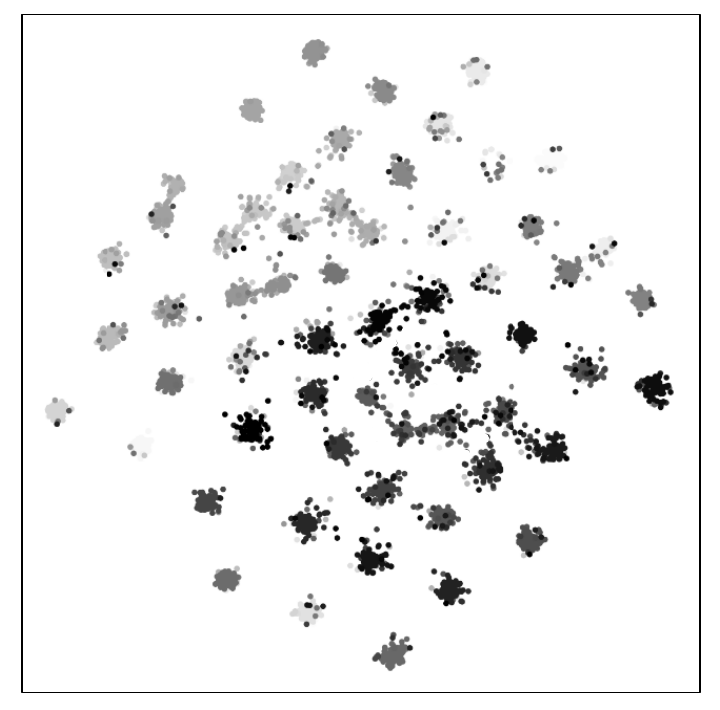}
    \label{fig:overlapc}}
    \subfloat[CE loss+$\delta $ (Base+Inc.)]{\includegraphics[width=0.4\linewidth]{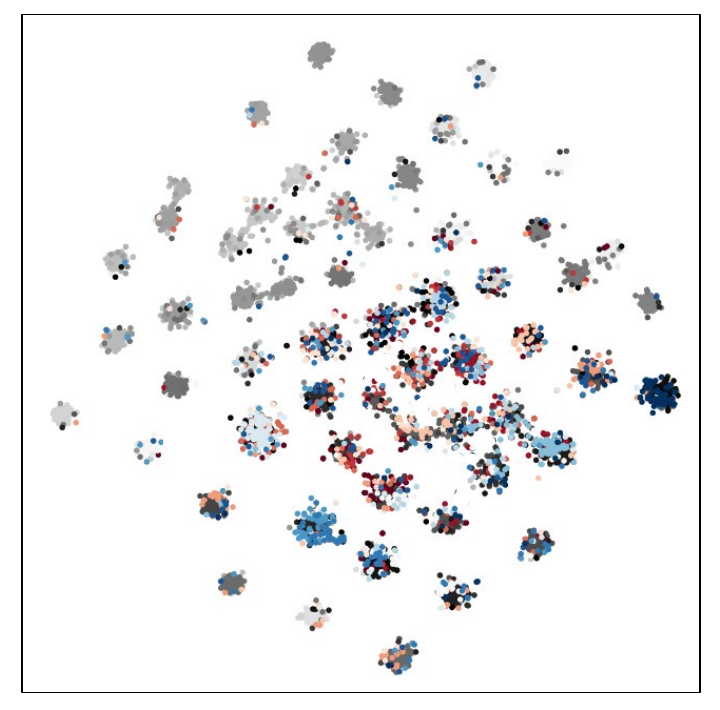}
    \label{fig:overlapd}}
    \caption{Visualization of the feature space with t-SNE on {\em mini}ImageNet test set. 
    The feature extractor is trained using only the base class samples. 
    The samples of $15$ incremental (Inc.) classes are scatteredly mapped to the base class positions.
    }
   \label{fig:overlap}
\end{figure}

In the FSCIL problem, the extremely limited novel class samples mean that traditional CIL training strategies cannot work effectively. Therefore, most FSCIL methods \cite{DBLP:conf/cvpr/ZhangSLZPX21, DBLP:conf/iclr/AkyurekAWA22, DBLP:journals/corr/abs-2312-05229} decouple the learning of representations and classifiers, train the whole model in the base session, then freeze feature extractor and only optimize classifiers in the incremental sessions. 
This strategy significantly alleviates {\em catastrophic forgetting} and {\em overfitting} in FSCIL, thus achieving significant results on the traditional evaluation criterion (i.e., overall classification accuracy), but {\em showing low accuracy for incremental classes (excluding base classes)}, as shown in \cref{fig:imba}.
Although recent methods\cite{DBLP:journals/corr/abs-2312-05229} have observed this phenomenon, it is simplistic to attribute its cause to the inaccuracy of incremental class classifiers.
However, by further visualizing the feature space, we further discover that the mapping results of the incremental class samples often overlap severely with the base classes, as shown in \cref{fig:overlap}.
That is, the incremental class samples may be inaccurately identified as they are sparsely mapped to the positions occupied by the base classes.

Recently, some methods\cite{DBLP:conf/cvpr/Zhao0XC0NF23, DBLP:journals/tip/JiHLPL23, DBLP:conf/iclr/KangYMHY23} readopt the traditional CIL training strategy, i.e., continuing to fine-tune the feature extractor with limited incremental class samples.
Although these methods slightly alleviate the above accuracy imbalance between base and incremental classes, {\em only the accuracy of newly arrived (current) incremental classes is relatively high, while the accuracy of past incremental classes is still low}, as shown in \cref{fig:imbb}. 
Moreover, due to the issue of overfitting, it is necessary to strictly control the range and degree of fine-tuning parameters, so there is still a large gap between the incremental class accuracy and the overall accuracy.

To achieve {\bf balanced} 
and {\bf effective} 
classification results throughout the incremental process, intuitively, all class samples
should be mapped to suitable positions in the feature space,
and all classifiers should gradually adapt to the incremental classes and data.
Thereafter, we could obtain an unbiased model, i.e., not overly biased towards certain classes due to sample quantity or arrival session, including base class and newly arrived (current) incremental class.

In this paper, we summarize the {\bf classification accuracy imbalance} phenomenon that is prevalent in FSCIL methods, systematically analyze the weaknesses and corresponding causes in existing methods, and abstract them into a unified {\bf model bias problem}. 
On this basis, we propose a method to significantly mitigate the model bias problem.
Specifically, 
based on the decoupling of the feature extractor and classifiers,
we employ approximate mixture distributions as classifiers, coupled with the semantic data, 
to stimulate the feature mapping ability and mitigate the feature extractor bias towards the mapping positions of base classes.
Furthermore, a separately dual-feature classification strategy is introduced
to preserve transferable features for future incremental classes while effectively training on base classes.
The transferable features are intelligently employed to optimize the classification results during inference, thereby mitigating the bias of the feature extractor towards discriminative features of base classes.
Finally, classifiers are continuously self-optimized based on the semantic distribution of incremental classes and the knowledge of novel samples 
to mitigate classifier bias throughout the entire incremental process. 
We summarize our method as `\textbf{S}timulation, \textbf{S}eparately, and \textbf{S}elf-optimizing', named SSS.
Our key contributions are summarized as follows:
\begin{itemize}
\item We summarize the prevalent classification accuracy imbalance phenomenon in FSCIL, systematically analyze the causes of this phenomenon, and abstract them into a unified model bias problem.
\item We propose the SSS method to mitigate model bias based on the analyses, and further improve it for the realistic scenario and fine-grained datasets.

\item 
Extensive experiments on benchmark datasets
show that our method 
significantly mitigates model bias and achieves state-of-the-art performance.


\end{itemize}

\section{Accuracy Imbalance in FSCIL}

\subsection{Definition and notations for FSCIL}
\label{sec:pro} 
In FSCIL, the model $f$ is trained on a sequence of datasets $\left\{\mathcal{D}_{train}^{t}\right\}_{t=0}^T$, where $\mathcal{D}_{train}^{t}=\left\{\left(\mathbf{x}_{i}, y_{i}\right)\right\}_{i}$ is the training set from session $t$ and $\mathbf{x}_{i}$ is a sample from class $y_{i} \in \mathcal{C}^{t}$. $\mathcal{C}^{t}$ is the label set of dataset $\mathcal{D}_{train}^{t}$. 
Usually, the training set $\mathcal{D}_{train}^{0}$ in the base session contains sufficient samples, and 
the training set $\mathcal{D}_{train}^{t} (t\ge 1)$ with the limited samples in each incremental session can be organized as $N$-way $K$-shot format, i.e., there are only $K$ samples for each of the $N$ classes from $\mathcal{C}^{t}$.
A model in each session $t$ can only access $\mathcal{D}_{train}^{t}$, but it needs to be tested on samples from all seen classes (i.e., $\mathcal{C}^0\cup\mathcal{C}^1\cdots\cup\mathcal{C}^t$).
To distinguish, we define classes in $\mathcal{C}^0$ as {\bf base classes}, in $\mathcal{C}^1\cdots\cup\mathcal{C}^t$ as {\bf incremental classes}, in $\mathcal{C}^1\cdots\cup\mathcal{C}^{t-1}$ as {\bf past incremental classes}, in $\mathcal{C}^{t}$ as {\bf current incremental classes} or {\bf novel classes}, and in $\mathcal{C}^0\cdots\cup\mathcal{C}^{t-1}$ as {\bf old classes}. 
The standard incremental learning paradigm strictly defines that each session has the same $N$ and $K$, and  $\mathcal{C}^{i} \cap \mathcal{C}^{j} = \varnothing$ for $i \ne j$. 

{\bf Training ($t=0$).} 
The model $f$ can be decomposed into feature extractor $g$ and classifiers $\eta$, to be specific,
\begin{equation}
  g(\mathbf{x}) = (\phi_1(\mathbf{x}),\phi_2(\mathbf{x}),\ldots,\phi_d(\mathbf{x}))^\top,\quad\phi_l\in\Phi.
  \label{eq:g}
\end{equation}
where $d$ is the vector dimension output by the feature extractor, $\phi_l $ is a feature mapping that maps $\mathbf{x}$ to $\mathbb{R}$ (including the process of global average pooling), so $\phi_l(\mathbf{x})$ can be referred to as a feature of $\mathbf{x}$. 
$g'(\mathbf{x}) = (\phi'_1(\mathbf{x}),\phi'_2(\mathbf{x}),\ldots,\phi'_d(\mathbf{x}))^\top$ denotes the $L_2$-normalized values of $g(\mathbf{x})$.

{\bf Inference.} 
Following \cite{DBLP:conf/nips/ShiCZZW21,DBLP:conf/aaai/MazumderSR21,DBLP:journals/tip/JiHLPL23}, 
we employ prototype classifiers rather than utilizing the trainable classifier $\eta $,
and utilize nearest class mean (NCM)\cite{DBLP:journals/pami/MensinkVPC13} algorithm for classification, which is defined as,
\begin{equation}
  y_{i}^\star=\underset{c\in {\textstyle \cup_{t=0}^{t'}} \mathcal{C}^{t}}{\operatorname{argmax}}\mathcal{S}(g(\mathbf{x}_i),P_c),
  \label{eq:cls}
\end{equation}
where $P_c$ indicates the prototype of class $c$ (the mean vector of all the training samples of class $c$), $\mathcal{S}(\cdot,\cdot)$ is used to measure the cosine similarity (If two sets of vectors are given, it calculates the cosine similarity for corresponding vectors and then takes the average), and $t'$ represents the current session.
For clarity, we define these classifiers generated from training samples as $h$, here $h=\{P_c| c\in {\textstyle \cup_{t=0}^{t'}} \mathcal{C}^{t}\}$.

{\bf Evaluation.}
Due to the irrationality of the evaluation metrics, most of the past methods ignored the issue of accuracy imbalance. 
Thus, we define {\bf Base acc.}, {\bf Inc. acc.}, {\bf CInc. acc.}, and {\bf PInc. acc.} based on the commonly used {\bf Overall acc.}, 
and define two accuracy ratios ({\bf Base/Inc.} and {\bf CInc./PInc.}) and {\bf BICP} in analyzing the accuracy imbalance phenomenon.
Please see \cref{sec:evaluation} for details.

\begin{figure*}[t]
  \centering
     \subfloat[Imbalance between base and incremental classes.]{\includegraphics[width=0.48\linewidth]{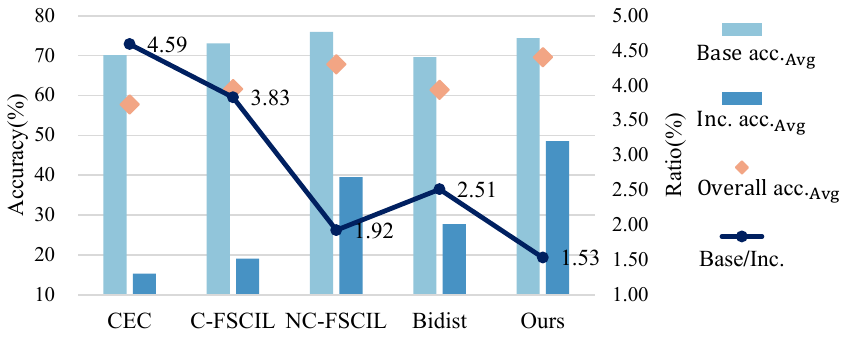}
    \label{fig:imba}}
    \hfill
     \subfloat[Imbalance between current and past incremental classes.]{\includegraphics[width=0.48\linewidth]{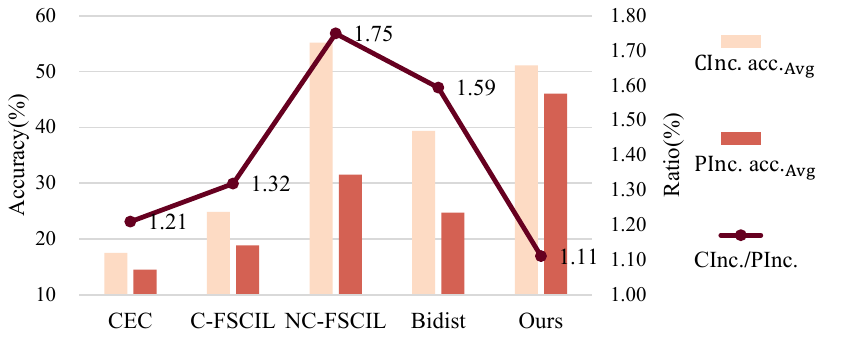}
    \label{fig:imbb}}
   \caption{Classification accuracy imbalance on {\em mini}ImageNet. 
   There is a serious accuracy imbalance between base and incremental classes in passive ways (e.g., CEC\cite{DBLP:conf/cvpr/ZhangSLZPX21}, C-FSCIL\cite{DBLP:conf/cvpr/HerscheKCBSR22}). Although active ways (e.g. NC-FSCIL\cite{DBLP:conf/iclr/YangYLLTT23}, Bidist\cite{DBLP:conf/cvpr/Zhao0XC0NF23}) somewhat alleviate the above imbalance, they cause a new accuracy imbalance between past and current incremental classes. Our method effectively alleviates the two types of accuracy imbalance and achieves the highest overall  accuracy ($\mathrm{Overall\ acc.}_{\mathrm{Avg.}}$).}
   \label{fig:imbalance}
\end{figure*}
\subsection{Observation and analyses}
\label{sec:ana}
A strong baseline\cite{DBLP:conf/cvpr/ZhangSLZPX21} for FSCIL is to train model $f$ with classification loss in the base session, and then freeze feature extractor $g$ in incremental sessions, 
thus significantly alleviating catastrophic forgetting and overfitting problems that FSCIL methods focus on.

Most FSCIL methods\cite{DBLP:conf/cvpr/ZhangSLZPX21,DBLP:conf/iclr/AkyurekAWA22, DBLP:conf/cvpr/HerscheKCBSR22, DBLP:journals/corr/abs-2312-05229} adopt the strong baseline, using the same feature extractor $g$ for both base classes and incremental classes.
However, using a feature extractor biased towards base classes for incremental classes seems to be just a {\bf passive way} to avoid catastrophic forgetting and overfitting.
In view of the neural collapse phenomenon\cite{Papyan2020PrevalenceON}, a feature extractor that only trained on base classes tends to map the feature vector of all samples to the same number of vertices as base classes in the feature space, and all vertices would form a simplex equiangular tight frame (ETF).
This weakens the mapping diversity and flexibility of feature extractor $g$, and greatly limits the feature space scope that can be mapped.
As shown in \cref{fig:overlapa,fig:overlapb}, we visualize the feature space of the {\em mini}ImageNet test set output from the feature extractor only trained on the base class samples using cross-entropy loss, 
and discover that the samples of $15$ incremental (Inc.) classes are scatteredly mapped to the base class positions. 
In addition, the base class classifier is more accurate, 
resulting in an extremely low average accuracy of incremental classes, 
so {\em the classification accuracy between base classes and incremental classes is imbalanced} (see \cref{fig:imba}).
Although using the base classifiers to improve the accuracy of incremental class classifiers can slightly alleviate this imbalance\cite{DBLP:journals/corr/abs-2312-05229}, it does not optimize the distribution of the base and incremental class samples in the feature space fundamentally.
Considering that boosting base class separation degree in base session may help fast generalization for novel classes\cite{DBLP:conf/cvpr/SongZSP0023}, 
we add a positive margin $\delta$ to cross-entropy loss as follows,
\begin{equation}
\mathcal{L} = \frac{1}{|\mathcal{D}_{train}^{0}|}\sum_{\mathbf{x_{i}} \in \mathcal{D}_{train}^{0}}\log\frac{e^{\eta_{y_i}^{\mathsf{T}}g(\mathbf{x_{i}})-\delta } }{\sum\limits_{j\ne y_i}e^{\eta_{j}^{\mathsf{T}}g(\mathbf{x_{i}})} +e^{\eta_{y_i}^{\mathsf{T}}g(\mathbf{x_{i}})-\delta }},
\end{equation}
and the feature space is visualized in \cref{fig:overlapc,fig:overlapd}. 
With clearer separation among base classes,
the incremental class samples would be more strictly mapped to the locations of base classes.
So simply boosting base class separation not only fails to alleviate the overlap between incremental class samples and base classes, but also exacerbates it.

Some recent methods\cite{DBLP:journals/tip/JiHLPL23, DBLP:journals/pami/YangLZLLJY23, DBLP:conf/cvpr/Zhao0XC0NF23} continue to fine-tune feature extractor with limited novel class samples, which slightly alleviates the classification accuracy imbalance between incremental classes and base classes. 
However, this {\bf active way} not only requires more complex network structures and training strategies, but also makes the feature vectors of old class samples gradually drift from their original classifiers.
To avoid feature drift, NC-FSCIL\cite{DBLP:conf/iclr/YangYLLTT23} predefines a number of fixed prototype classifiers, and only fine-tunes a projection layer between the backbone and classifiers when novel classes arrive. 
Considering the projection layer as a deep component of the feature extractor, this method essentially involves fine-tuning the feature extractor.
When updating the projection layer, the model has access solely to current incremental class samples and the mean intermediate feature of old classes. 
Therefore, this active way inevitably leads to catastrophic forgetting of past incremental classes again, resulting in {\em an accuracy imbalance between past incremental classes and current incremental classes} (see \cref{fig:imbb}).
Besides, due to the constraint of overfitting, the range and degree of fine-tuning parameters need to be strictly controlled, so there is still an accuracy imbalance between base classes and incremental classes.

\subsection{Solutions: mitigate model bias}
\label{sec:solu}

It is known that fine-tuning the feature extractor based on limited novel class samples would make the model biased towards novel classes again, thus causing the accuracy imbalance between past and current incremental classes. 
{\em But if the feature extractor is not fine-tuned in incremental sessions, 
how can it balance the accuracy between base classes and future incremental classes?}



{\bf Enhancing randomness in the feature level of feature mappings.}
As analyzed in \cref{sec:ana}, the key to improving the accuracy of future incremental classes is to mitigate the bias of the feature extractor towards the mapping positions of base classes.
Although FACT\cite{DBLP:conf/cvpr/0001WYMPZ22} pre-assigns prototypes to 
reserve space for incremental classes, 
feature mappings may not necessarily possess the ability to map samples into these positions.
In addition, SAVC\cite{DBLP:conf/cvpr/SongZSP0023} generates virtual class samples
to act as placeholders for incremental classes, but it still aggregates all virtual class results to identify base classes, i.e., 
the locations of placeholders are still utilized by base classes.
Therefore, {\em not only the diversity and flexibility of the feature mapping results should be ensured, but also the mappable positions should not be all occupied by base classes}. 
This mitigates the bias of the feature extractor for the mapping position.

{\bf Reserving the determinism in the semantic level of feature mappings.}
To achieve effective classification, incremental classes should be mapped into clusters according to semantics, rather than being scatteredly mapped to positions without base classes.
However, 
the feature extractor tends to retain {\bf class-specific discriminative features} in \cref{def:c} that are highly correlated with classes supervised by the loss function, while losing {\bf transferable features} in \cref{def:t} that may be used to identify future incremental classes but interfere with current classification accuracy. 
Although ALICE\cite{DBLP:conf/eccv/PengZWLL22} directly uses the projector head of SimCLR\cite{DBLP:conf/icml/ChenK0H20} to avoid feature extractor overfitting base classes, it largely sacrifices the accuracy of base classes.
Thus, {\em it is necessary to preserve transferable features for future incremental classes while ensuring effective training on base classes,
and allowing transferable features to optimize the classification process}.
This mitigates the bias of the feature extractor for feature retaining.



{\bf Achieving the joint prosperity of all classifiers besides the trade-off between base and incremental classifiers.}
Given the class sample imbalance for classifier construction and learning the continuously complexifying classification task during the dynamic incremental process, {\em it is crucial to persistently optimize existing classifiers based on the semantic distribution of classes and the knowledge of novel samples throughout the incremental process}. 
This mitigates the bias of different class classifiers in accuracy.

\section{Methodology}
According to analyses in \cref{sec:solu}, we propose the SSS method for model bias mitigating in the FSCIL task. 
Specifically, for feature extractor bias mitigating, we design mapping ability stimulation and separately dual-feature classification in \cref{sec:f,sec:mlp}; for classifier bias mitigating, we propose self-optimizing classifiers in \cref{sec:cls}.


\subsection{Mapping ability stimulation}
\label{sec:f}
\begin{figure}[t]
  \centering
   \includegraphics[width=1\linewidth]{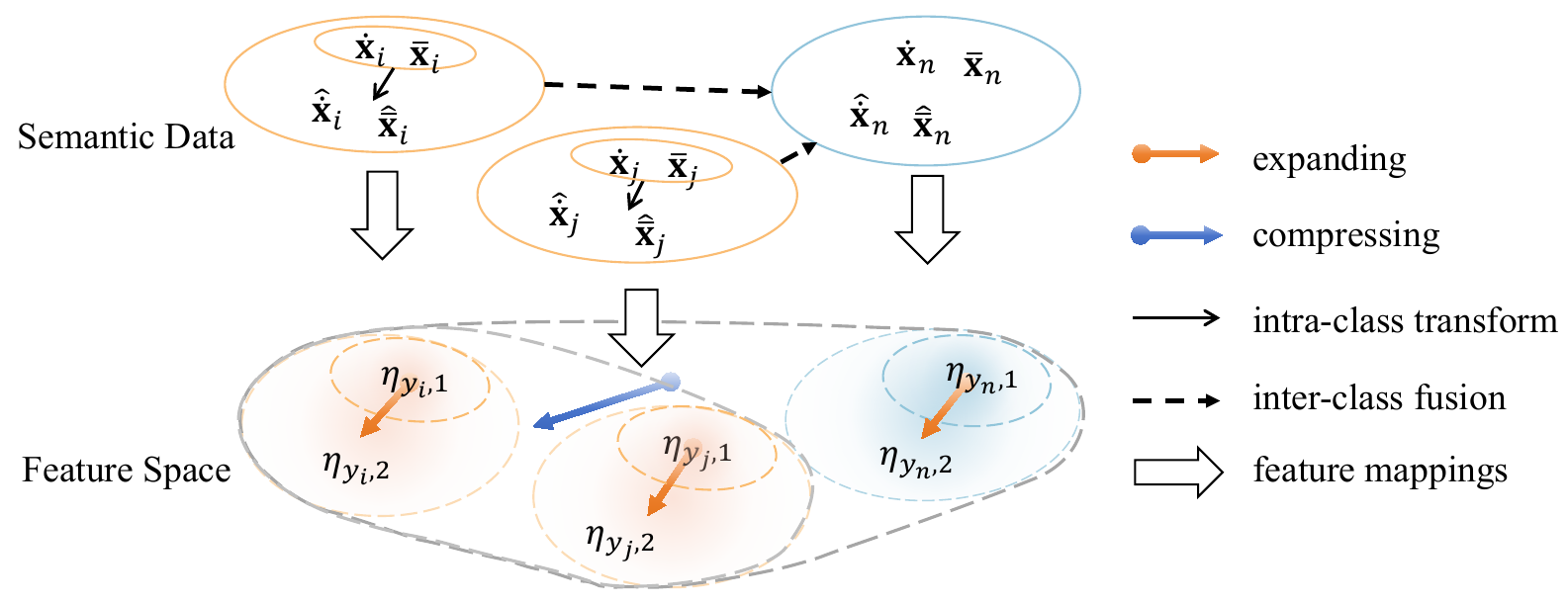}
   \caption{The component to stimulate mapping ability, including intra-class transform and inter-class fusion, to expand overall mappable space and compress the feature space occupied by base classes for future incremental classes.}
   \label{fig:stimulate}
\end{figure}

In order to stimulate mapping ability of the feature extractor and obtain diverse and flexible mapping results, we improve the training process in \cref{sec:pro} by introducing approximate mixture distribution based classifiers to compress and expand feature space.
Thereafter, the optimization goal of the feature extractor is to map all samples of different classes to the surplus mixture distributions.
As shown in \cref{fig:stimulate}, surplus approximate mixture distribution based classifiers and semantic data serve the purpose of {\em expanding} overall mappable space while {\em compressing} the feature space occupied by base classes, i.e., achieving diversity and unoccupied feature mappings for future incremental classes,
thereby mitigating the bias towards the mapping positions of base classes.

Concretely, a mixture distribution consists of two components with equal weights (here it is simply represented by two trainable classification vectors $\eta_{c,j}(j=1,2)$), and there exist far more mixture distributions than the number of base classes. 
To fit the distributions, intra-class image transformation and inter-class fusion are adopted as follows,
\begin{equation}
\begin{bmatrix} \hat{\dot{\mathbf{x}}}_{i} \\ \hat{\bar{\mathbf{x}}}_{i} \end{bmatrix} = \mathcal{T} \left(\begin{bmatrix} \dot{\mathbf{x}}_i \\ \bar{\mathbf{x}}_i \end{bmatrix}\right),
\end{equation}
\begin{equation}
\begin{bmatrix} \dot{\mathbf{x}}_n & \bar{\mathbf{x}}_{n}\\ \hat{\dot{\mathbf{x}}}_{n}& \hat{\bar{\mathbf{x}}}_{n} \end{bmatrix} = \lambda \begin{bmatrix} \dot{\mathbf{x}}_i & \bar{\mathbf{x}}_{i}\\ \hat{\dot{\mathbf{x}}}_{i}& \hat{\bar{\mathbf{x}}}_{i} \end{bmatrix} + (1-\lambda)\begin{bmatrix} \dot{\mathbf{x}}_j & \bar{\mathbf{x}}_{j}\\ \hat{\dot{\mathbf{x}}}_{j}& \hat{\bar{\mathbf{x}}}_{j} \end{bmatrix},j \neq i,
\label{eq:lamda}
\end{equation}
where each image $\mathbf{x}_i$ within each class is randomly preprocessed to two images $\dot{\mathbf{x}}_i$ and $\bar{\mathbf{x}}_i$ corresponding to one target component $\eta_{y_{i},1}$;
and then they are transformed into $\hat{\dot{\mathbf{x}}}_{i}$ and $\hat{\bar{\mathbf{x}}}_{i}$ corresponding to another target component $\eta_{y_{i},2}$
($\mathcal{T}$ represents the vertical flip operation).
Thereafter, the random cross-class images $\mathbf{x}_i$ and $\mathbf{x}_j$ are fused into $\mathbf{x}_n$ to fit a surplus mixture distribution $\eta_{y_{n}}$ ($\lambda$ is sampled from Beta distribution).

\subsection{Separately dual-feature classification}
\label{sec:mlp}
Even if samples can be mapped to positions that are not occupied by base classes, 
the feature extractor directly involved in base training inevitably tends to focus on {\em class-specific discriminative features} for base training classes. 
Consequently, despite the presence of diverse and flexible feature mapping results, 
the incremental classes may not necessarily form clusters in the feature space without effective discriminative features. 
Hence, we introduce a separately dual-feature classification strategy in \cref{alg:mlp},
intelligently combining the {\em transferable features} maintained during training with the final {\em class-specific discriminative features} to optimize the classification results during inference.
\begin{figure}[t]
 \begin{algorithm}[H]
  \caption{Separately dual-feature classification strategy.}
    \label{alg:mlp}
    \begin{algorithmic}[1]
 
    \REQUIRE Training set $\mathcal{D}_{train}^{t}$ and test set $\mathcal{D}_{test}^{t}$ in session $t$, feature extractor $\tilde{g}$ (including $g$ and $SR$), classifiers $h$ and $\tilde{h}$ for old classes.
    \ENSURE Test set classification results.
 
    \STATE $\mathcal{X}$, $\tilde{\mathcal{X}}$$\gets$ Extract transferable and class-specific discriminative features from $\mathcal{D}_{train}^{t}$ using $g$ and $\tilde{g}$;
 
    \STATE $h_{novel}$,$\tilde{h}_{novel}$$\gets$ Generate novel classifiers from $\mathcal{X}$, $\tilde{\mathcal{X}}$;
    \STATE $h$, $\tilde{h}$ $\gets$ $ h \cup h_{novel}$, $\tilde{h} \cup \tilde{h}_{novel}$
    \FOR{each $\mathbf{x}_{i}$ in $\mathcal{D}_{test}^{t}$} 
    \STATE $y_{i}$$ \gets$ classify $g(\mathbf{x}_{i})$ using $h$ as \cref{eq:cls};
    \IF {$y_{i}$ not in $\mathcal{C}_{0}$}
    \STATE Output the classification result $y_{i}$;
    \ELSE
        \STATE $\tilde{y}_{i}$$ \gets$ classify $\tilde{g}(\mathbf{x}_{i})$ using $\tilde{h}$ as \cref{eq:cls};
        \STATE Output the classification result $\tilde{y}_{i}$;
      \ENDIF
    \ENDFOR
   \end{algorithmic}
 \end{algorithm}
\end{figure}

Specifically, given test image $\mathbf{x}_{i}$, its transferable feature $g (\mathbf{x}_{i})$ and class-specific discriminative feature $\tilde{g} (\mathbf{x}_i)$ are obtained separately. 
Thereafter, classifiers $h$ are first generated based on transferable features as introduced in the inference process in \cref{sec:pro} to achieve a preliminary, coarse classification. 
If $\mathbf{x}_{i}$ is categorized as a base class sample, it will be further verified with its class-specific discriminative features $\tilde{g} (\mathbf{x}_i)$ and corresponding classifiers $\tilde{h}$. Such two-step strategy can facilitate more precise recognition of novel class features similar to discriminative features of the base training classes, as well as those that may be ignored during base class training.
Besides, the boundaries among base classes become clearer, ensuring the accuracy of the base classes as well.
In order to preserve the transferability of features, one possible approach is introducing an isolating module between target features and the classification task, to mitigate feature extractor bias towards the current task without compromising training effectiveness. In this paper, a selection and reorganization (SR) module is added after the original feature extractor to form a new feature extractor for classification vectors generation,
\begin{equation}
  \tilde{g} (\mathbf{x}) = SR(g(\mathbf{x})),
  \label{eq:sr}
\end{equation}
where $SR$ denotes a block that consists of two fully connected layers and a ReLU activation function.
Guided by the loss function during training, $\tilde{g} (\mathbf{x})$, i.e., class-specific discriminative features, becomes more discriminative for recognizing base training classes through the process of selecting and reorganization. 
Meanwhile, $g$ is encouraged to extract richer $g(\mathbf{x})$, i.e., transferable features, for subsequent selection and reorganization.
The subsequent sections consistently employ '$\sim$' to distinguish symbols associated with the two types of features mentioned above.
To align with the training phase,
classifiers $h=\{[P_{c,1}, P_{c,2}]| c\in {\textstyle \cup_{t=0}^{t'}} \mathcal{C}^{t}\}$ (generated from the training samples and their vertically flipped counterpart) in the inference phase, and the test sample feature $g (\mathbf{x}_i)=[g (\mathbf{x}_i), g (\hat{\mathbf{x}}_i)]$.

\subsection{Self-optimizing classifiers}
\label{sec:cls}

To mitigate classifier bias throughout the dynamic incremental process, all existing classifiers should undergo self-optimization to adapt to the new sample distribution in the feature space. 
In this section, we first individually consider base classifiers and then attempt to propose a holistic scheme for all classifiers.

\begin{figure}[t]
  \centering
   \includegraphics[width=0.9\linewidth]{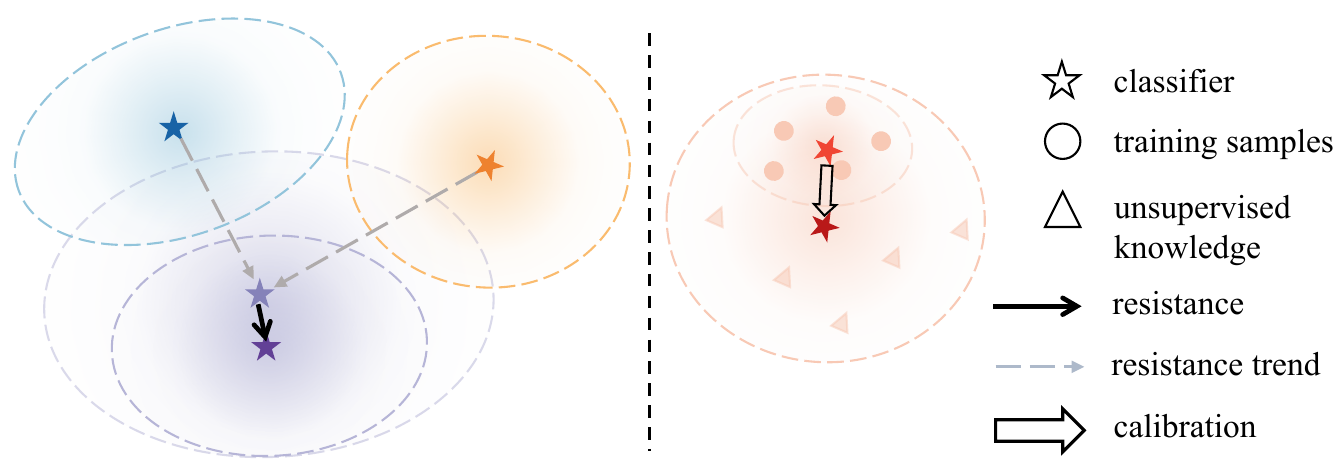}
   \caption{The components to self-optimizing classifiers,
   i.e., resisting base class classifiers using the incremental classes and continuously calibrating existing classifiers based on encountered samples.}
   \label{fig:opt}
\end{figure}

As introduced in \cref{sec:mlp}, the classifiers $\tilde{h}$ aim to distinguish base class samples and samples of incremental classes that are similar to base classes in discriminative features. 
To mitigate the classification results bias towards base classes, we propose the idea of \textit{Resistance} to optimize base class classifiers according to the semantic distribution of incremental classes as illustrated in the left figure of \cref{fig:opt}. 
Given that continuous optimization based on incremental classes would lead to the reintroduction of forgetting issues for base classes, we design this resistance operation to function as a one-time process only before each inference. Specifically, 
to avoid excessive resistance, 
prototypes $\tilde{P}_{c,j}$ stay as the initial state during the self-optimization throughout incremental sessions, and will be updated only for inference,

\begin{equation}
    \tilde{P}'_{c,j}  = \tilde{P}_{c,j} - \gamma \cdot \frac{\Delta _{c,j}}{\left \|  \Delta _{c,j}\right \|}, c \in \mathcal{C}^{0}, j=1,2,
  \label{eq:re1}
\end{equation}
\begin{equation}
    \Delta _{c,j}  \gets \Delta _{c,j} + \sum_{i \in \mathcal{C}^{t}} \max \left(\mathcal{S} (\tilde{P}_{c,j},\tilde{P}_{i,j}),0\right ) \cdot  \frac{\tilde{P}_{i,j}}{\left \|  \tilde{P}_{i,j}\right \|},  
  \label{eq:re2}
\end{equation}
where 
$\Delta _{c,j} $ summarizes the main directions in which incremental classes exist, and is continuously updated in each session; 
$\gamma$ is a random number used to amplify the resistance,
and $\max(x,0)$ is used to select the novel class prototype $\tilde{P}_{i,j}$ that has a cosine similarity greater than 0 with the base class prototype $\tilde{P}_{c,j}$. 


\subsubsection{Self-optimizing in realistic scenario} 
\label{sec:realistc}
For convenient experiments and evaluation, previous FSCIL methods adhere to the same problem setting as CIL, i.e., novel samples for training invariably exclude any old class samples. The sections above in this paper also conventionally follow this setting. 
However, this rigid and strict setting, coupled with the extremely limited data in the incremental sessions of FSCIL, greatly restricts the potential for optimizing the classifiers during the incremental session. This limitation renders existing methods incapable of proposing comprehensive optimization solutions for the classifier. Consequently, whether it is moving the base class classifiers (i.e., our {\em Resistance} idea) or moving the incremental class classifiers \cite{DBLP:journals/corr/abs-2312-05229}, fundamentally, it is a compromise that sacrifices the accuracy of some classes to improve the accuracy of other classes.

In fact, 
unlike traditional static models that follow a strict division of training and inference phases after initial training, incremental learning methods operate in dynamic environments. There is no strict division between training and inference phases in practical applications, implying that, theoretically, the model can leverage all the data it has encountered (both during training or inference) up to a certain moment to optimize its performance on subsequent data. As a result, the conventional settings mentioned above not only fail to accurately represent FSCIL but also hinder the methods from addressing the real FSCIL task.

Therefore, building upon the superior performance achieved in the standard FSCIL experimental setting, our method takes the first step to truly consider FSCIL as an open-world dynamic task. 

Specifically, we propose the idea of {\em Calibration} to fully utilize all the data available, including few-shot incremental class training data and unlabeled test samples, to achieve joint prosperity of all classifiers.
That is, the prototype classifiers $h$
are calibrated by taking the weighted average of the prototype $P_{c,j}$ and the feature vectors of the unlabeled samples that are recognized as belonging to the prototype class $c$,
\begin{equation}
  P_{c,j}  \gets (1-\alpha) \cdot P_{c,j}+\alpha \cdot avg(\mathcal{X}_{test}), j=1,2,
  \label{eq:ca1}
\end{equation}
\begin{equation}
\forall g(\mathbf{x}_{i}) \in \mathcal{X}_{test},  \mathcal{S}(g(\mathbf{x}_{i}),P_{c,j})>r.
\label{eq:ca2}
\end{equation}
where 
$\mathcal{X}_{test}$ is the set of at most $R$ feature vectors of unlabeled test set samples that have a cosine similarity higher than a threshold $r$ with the prototype $P_{c,j}$, $avg(\cdot)$ calculates the average vector of the vector set $\mathcal{X}_{test}$, and $\alpha$ is used to control the degree of calibration. 
If old class labeled samples are encountered in the incremental sessions, they can also be effectively utilized as \cref{eq:ca1}, and then $\alpha$ can be directly calculated based on the ratio between the number of newly added samples and the past sample quantity.

\subsubsection{Self-optimizing for fine-grained dataset} 
Considering the importance of detail differences and discriminative information for fine-grained data, we additionally propose a classifiers optimization scheme based on Bayesian Gaussian mixture model (BGMM) to provide more fine-grained description. See \cref{sec:BGMM} for details.



\section{Experiments}
\subsection{Experimental setup}
{\bf Datasets.}
Following \cite{DBLP:conf/cvpr/TaoHCDWG20}, we conduct experiments on three datasets: {\em mini}ImageNet\cite{DBLP:journals/ijcv/RussakovskyDSKS15}, CIFAR100\cite{Krizhevsky2009LearningML}, and CUB200\cite{Wah2011TheCB}.
Please see \cref{sec:setup} for details. 

\noindent{\bf Implementation Details.} 
Following \cite{DBLP:conf/cvpr/TaoHCDWG20}, we employ ResNet18\cite{DBLP:conf/cvpr/HeZRS16} as the backbone, and the network for CUB200 is initialized by ImageNet\cite{DBLP:conf/cvpr/DengDSLL009} pre-trained parameters. 
The results of the comparative methods that are not reported in their papers are reproduced by their publicly available source code.
Please see \cref{sec:setup} for details. 

\begin{table*}[htbp]
  \centering
  \caption{Comparison with SOTAs on {\em mini}ImageNet dataset. 
Due to space limitations, only the Inc. acc. and PInc. acc. on sessions 2, 5, and 8, and the average value of Overall acc. are presented.
Please refer to \cref{sec:results} for detailed results on three datasets.
    }
    \begin{tabular}{lccc|ccccc}
    \toprule
    \multicolumn{1}{c}{\multirow{2}[4]{*}{Method}} & \multicolumn{3}{c|}{Inc. acc.(\%)} & \multicolumn{3}{c}{PInc. acc.(\%)} & \multirow{2}[4]{*}{$\mathrm{Overall\ acc.}_{\mathrm{Avg.}}$} & \multirow{2}[4]{*}{BIPC} \\
\cmidrule{2-7}          & 2     & 5     & 8     & 2     & 5     & 8     &       &  \\
    \midrule
    CEC\cite{DBLP:conf/cvpr/ZhangSLZPX21}   & 17.00  & 14.32  & 14.88  & 13.80  & 15.05  & 13.97  & 57.75  & 2.90  \\
    FACT\cite{DBLP:conf/cvpr/0001WYMPZ22}  & 14.40  & 13.64  & 13.20  & 14.20  & 13.55  & 12.34  & 59.88  & 3.21  \\
    C-FSCIL\cite{DBLP:conf/cvpr/HerscheKCBSR22} & 12.40  & 23.40  & 25.95  & 6.40  & 20.80  & 26.09  & 61.61  & 2.56  \\
    TEEN\cite{DBLP:journals/corr/abs-2312-05229}  & 35.60  & 29.96  & 29.35  & 35.80  & 31.55  & 27.57  & 61.44  & 1.62  \\
    Bidist\cite{DBLP:conf/cvpr/Zhao0XC0NF23} & 30.30  & 28.44  & 25.62  & 25.80  & 24.50  & 23.74  & 61.42  & 2.05  \\
    SAVC\cite{DBLP:conf/cvpr/SongZSP0023}  & 30.30  & 27.32  & 26.58  & 30.40  & 28.15  & 25.40  & 67.05  & 1.92  \\
    NC-FSCIL\cite{DBLP:conf/iclr/YangYLLTT23} & 48.10  & 34.88  & 31.33  & 33.00  & 30.15  & 27.74  & 67.82  & 1.84  \\
    \midrule
    Ours(Prototype) & \textbf{54.20} & \textbf{44.64} & \textbf{45.35} & \textbf{54.20} & \textbf{45.20} & \textbf{43.91} & \textbf{69.60} & 1.32  \\ 
    Ours(BGMM) & 54.00  & 44.16  & 44.23  & 54.00  & 45.00  & 42.57  & 69.50  & \textbf{1.33} \\
    \bottomrule
    \end{tabular}%
  \label{tab:comp}%
\end{table*}%

\begin{figure}[!t]
  \centering
   \includegraphics[width=0.96\linewidth]{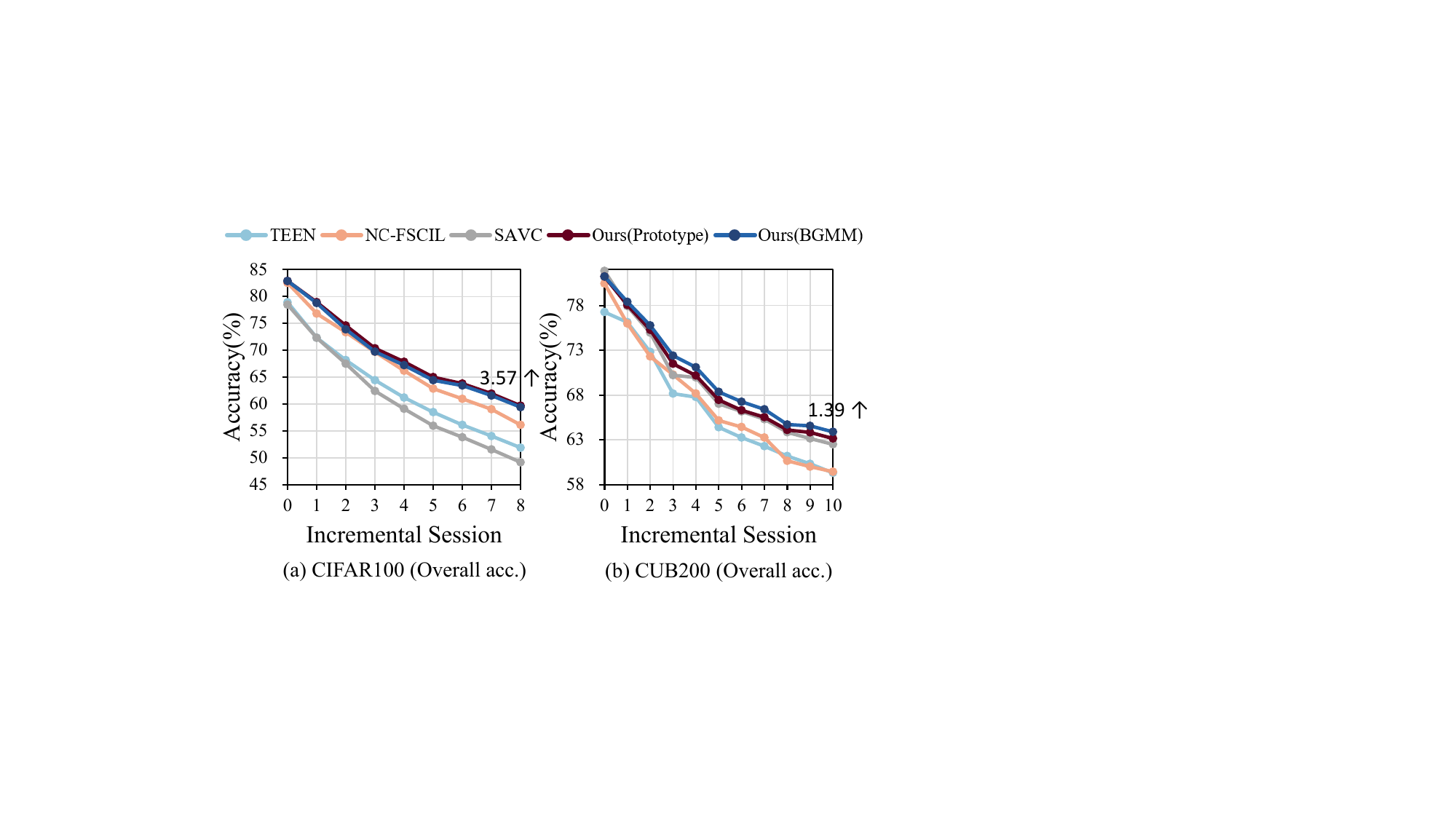}
   \caption{Comparison with SOTAs on CIFAR100 and CUB200 datasets in terms of the overall accuracy.}
   \label{fig:comp}
\end{figure}

\subsection{Comparisons with state-of-the-arts}
We compare our method with recent SOTA FSCIL methods on three widely used datasets.
As presented in \cref{tab:comp}, our method significantly outperforms the SOTAs in terms of the commonly used overall accuracy (Overall acc.),
as well as the incremental class accuracy (Inc. acc.) and past incremental class accuracy (PInc. acc.), which are poorly performed by existing FSCIL methods. 
Importantly, our method achieves the lowest BICP, that is, the lowest degree of accuracy imbalance.

After effectively mitigating the accuracy imbalance caused by the model bias problem, our method reduces the model’s forgetting rate in a practical sense, thus achieving excellent overall performance in the incremental sessions.
Specifically, 
our method shows a mere 0.36\% superiority over NC-FSCIL\nocite{DBLP:conf/iclr/KangYMHY23} in the base session, whereas in the last session, it outperforms NC-FSCIL by 3.57\% on CIFAR100 (see \cref{fig:comp}(a)). 
In addition, 
the performance of BGMM based classifers is significantly better on the fine-grained dataset CUB200 (see \cref{fig:comp}(b) and \cref{tab:cub1}). 

\begin{table*}[htbp]
  \centering
  \caption{ Ablation studies of our proposed method on prototype based classifiers on CIFAR100 dataset.}
    \begin{tabular}{ccc|cc|cccccccc|c}
    \toprule
    \multicolumn{3}{c|}{ Feature extractor} & \multicolumn{2}{c|}{Classifiers} & \multicolumn{8}{c|}{Inc. acc. (\%)}                           & \multirow{2}[4]{*}{Base/Inc.} \\
\cmidrule{1-13}    \textbf{Intra} & \textbf{Inter} & \textbf{SR} & \textbf{R} & \textbf{C} & 1     & 2     & 3     & 4     & 5     & 6     & 7     & 8     &   \\
    \midrule
          &       &       &       &       & 22.00  & 17.80  & 15.73  & 14.50  & 13.80  & 13.30  & 12.74  & 13.13  & 4.92  \\
    $\surd$     &       &       &       &       & 30.40  & 19.90  & 17.60  & 16.10  & 16.48  & 16.27  & 15.71  & 16.00  & 4.19  \\
    $\surd$     & $\surd$     &       &       &       & 36.20  & 30.50  & 23.73  & 22.10  & 21.12  & 19.83  & 18.49  & 17.90  & 3.33  \\
    $\surd$     & $\surd$     & $\surd$     &       &       & 50.00  & 44.70  & 37.47  & 38.05  & 36.96  & 39.13  & 38.20  & 36.65  & 1.94  \\
    \midrule
    $\surd$     & $\surd$     & $\surd$     & $\surd$    &       & 63.40  & 61.70  & 51.00  & 50.00  & 49.04  & 50.47  & 48.17  & 46.37  & 1.40  \\
    $\surd$     & $\surd$     & $\surd$    & $\surd$    & $\surd$    & \textbf{63.80} & \textbf{61.90} & \textbf{51.47} & \textbf{50.60} & \textbf{49.72} & \textbf{51.13} & \textbf{48.80} & \textbf{47.15} & \textbf{1.38} \\
    \bottomrule
    \end{tabular}%
  \label{tab:abl}%
\end{table*}%

\subsection{Ablation studies and analyses}
To analyze the role of different components in mitigating model bias, we conduct ablation studies
on CIFAR100 dataset. The results are reported in \cref{tab:abl}.

\begin{figure}[!t]
  \centering
     \subfloat[Base]{\includegraphics[width=0.47\linewidth]{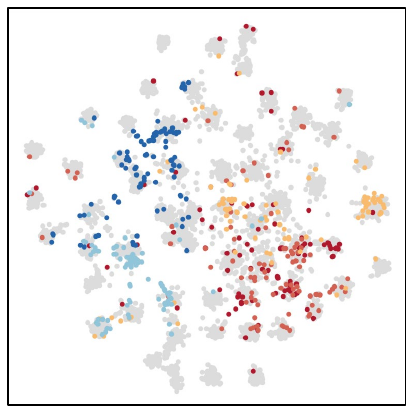}
    \label{fig:visa}}
    \subfloat[Base+Intra]{\includegraphics[width=0.47\linewidth]{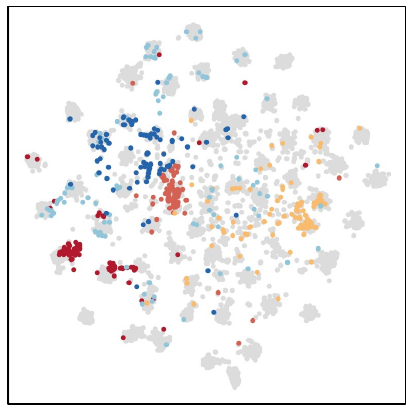}
    \label{fig:visb}}
   
    \subfloat[Base+Intra+Inter]{\includegraphics[width=0.47\linewidth]{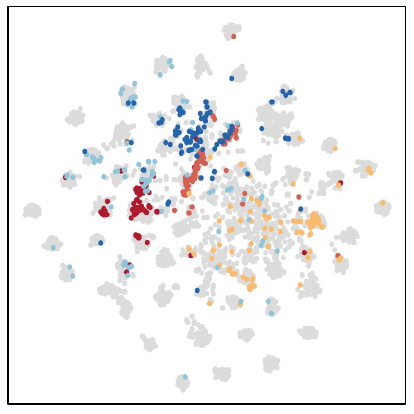}
    \label{fig:visc}}
    \subfloat[Base+Intra+Inter+SR($g(\mathbf{x})$)]{\includegraphics[width=0.47\linewidth]{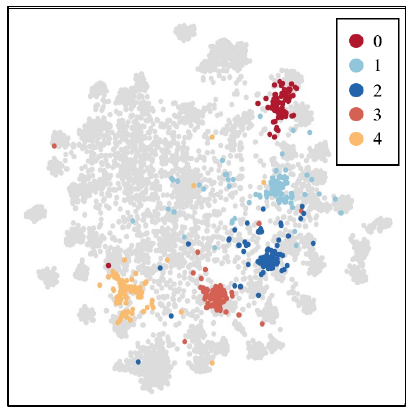}
    \label{fig:visd}}
  \caption{Visualization of feature space with t-SNE\nocite{Maaten2008VisualizingDU} on CIFAR100 test set. Gray represents base classes and other colors represent incremental classes. Please see \cref{sec:vis1} for detailed analyses.}
  \label{fig:vis1}
\end{figure}

Firstly, Intra-class transformation (\textbf{Intra}) and inter-class fusion (\textbf{Inter}) preliminarily alleviate the accuracy imbalance between base and incremental classes.
We showcase Feature Mapping Occupancy ({\bf FMO}) in \cref{def:fmp} for base class samples (see \cref{sec:fmo}). 
Introducing Intra initially reduces FMO, essentially expanding the feature space while maintaining the feature dimensions unchanged, 
and Inter further reduces FMO, 
i.e., compressing the feature space occupied by base classes.
Visualization of feature space demonstrates that our method can effectively mitigate the mapping bias towards the base class positions (see \cref{fig:visa,fig:visb,fig:visc}).

\begin{figure}[t]
  \centering
  \subfloat[$g(\mathbf{x})$]{\includegraphics[width=0.45\linewidth]{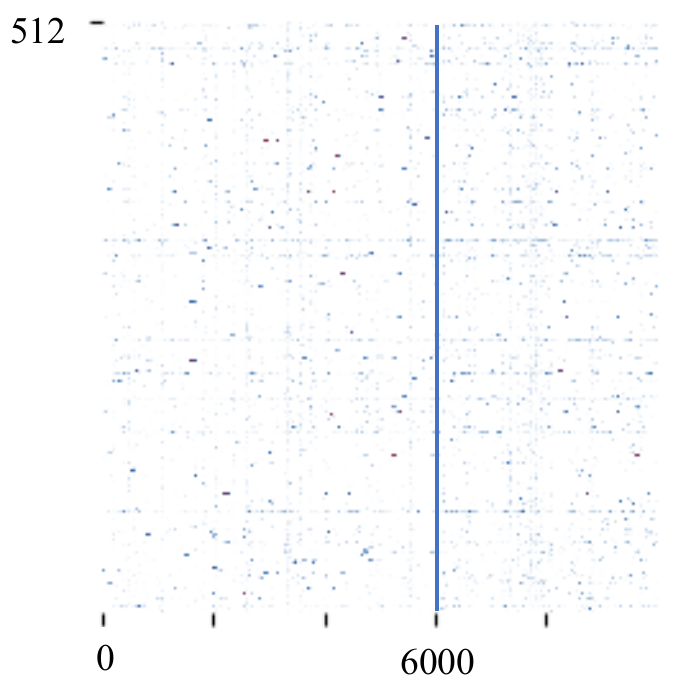}
    \label{fig:ha}}
    \subfloat[$g(\mathbf{x})(+SR)$]{\includegraphics[width=0.54\linewidth]{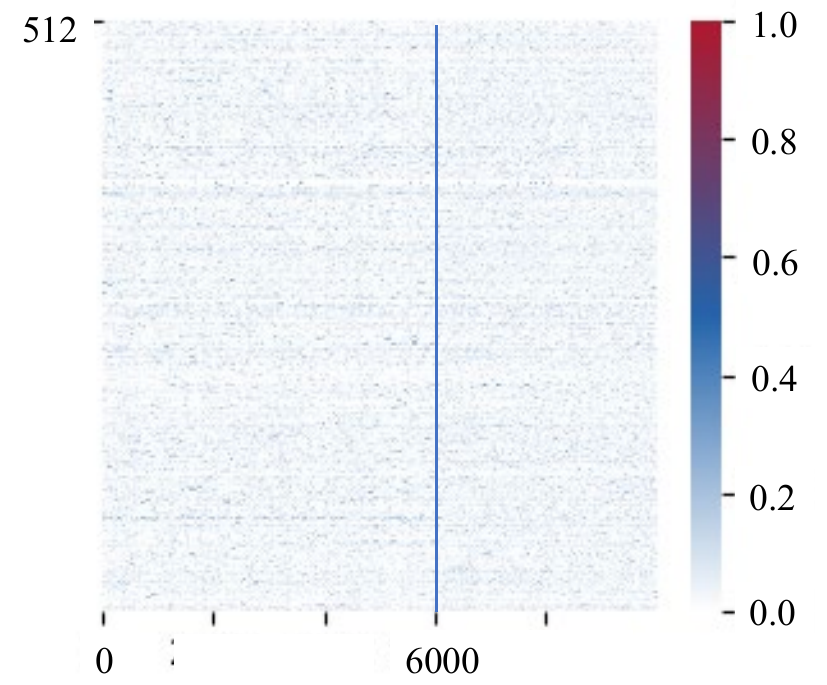}
    \label{fig:hb}}
   \caption{Comparison of heatmaps of $L_2$ normalized feature vectors on whether adding $SR$ after $g$ on CIFAR100 test set. The higher the value, the stronger the discriminative power of features.
   }
   \label{fig:heatmap}
\end{figure}

Further, the separately dual-feature classification with the SR module (\textbf{SR}) further significantly alleviates this imbalance.
We visualize the $L_2$ normalized feature vectors on CIFAR100 test set in \cref{fig:heatmap}, which demonstrate that the separately feature extraction idea with the SR module added after $g$ can indeed stimulate $g(\mathbf{x})$ to learn and retain more transferable features for incremental classes (see \cref{sec:fig7} for details).
Transferable features $g(\mathbf{x})$ ensures that incremental classes can cluster effectively based on semantic categories, but they blur the boundaries among base classes and between base and incremental classes (see \cref{fig:visd}). 
Therefore, we propose the separately dual-feature classification to intelligently combine transferable features $g (\mathbf{x})$ with class-specific discriminative features $\tilde{g} (\mathbf{x})$.
In addition, we compare the performance of different combined ways on CIFAR100 dataset in \cref{tab:2f1}. It demonstrates that our separately dual-feature classification idea achieves the best performance. 

\begin{table}[htbp]
  \centering
  \caption{$\mathrm{Inc.\ acc.}_{\mathrm{Avg.}}$(the first line) and $\mathrm{Overall\ acc.}_{\mathrm{Avg.}}$(the second line) of our separately dual-feature classification comparing to other ways to utilizing transferable features $g(\mathbf{x})$ and class-specific discriminative features $\tilde{g}(\mathbf{x})$ on CIFAR100 dataset. Please see \cref{sec:tab3} for detailed results and analyses.}
    \begin{tabular}{cccccc}
    \toprule
     $g(\mathbf{x})$  & $\tilde{g}(\mathbf{x})$ & \textbf{Pre}   & \textbf{Post}  & \textbf{AD} & Ours \\
    \midrule
    35.06  & 20.40  & 20.92  & 18.99  & 19.63  & \textbf{40.15} \\
    67.97  & 66.29  & 66.44  & 65.02  & 67.99  & \textbf{69.52} \\
    \bottomrule
    \end{tabular}%
  \label{tab:2f1}%
\end{table}%





It is worth noting that only mitigating feature extractor bias towards the base classes significantly improves the incremental class accuracy (and also achieves the obviously highest Inc. acc. and the lowest Base/Inc. compared to existing methods, see \cref{sec:results} for the data of other methods).

On this basis, resisting (\textbf{R}) and calibration (\textbf{C}) further alleviate the accuracy imbalance between base and incremental classes.
As mentioned in \cref{sec:realistc}, based on a more realistic consideration, \textbf{C} achieves more obvious optimization for classifiers.
Since the open-set performance of a classifier can be improved by enhancing its closed-set accuracy\cite{DBLP:conf/iclr/Vaze0VZ22}, through this mutually reinforcing process, the classifier would gradually approach the optimal from a holistic perspective.




\section{Conclusion}
In this paper, we summarize the classification accuracy imbalance phenomenon that is prevalent in FSCIL methods, then find the causes of this phenomenon and abstract them into a unified model bias problem.
Based on the analyses, we propose a method (SSS) to mitigate the problem.
Extensive experiments show that our method significantly mitigates the model bias problem and achieves state-of-the-art performance.

\bibliography{example_paper}
\bibliographystyle{icml2024}

\newpage
\appendix
\onecolumn

\section{Evaluation Metrics}
\label{sec:evaluation}
FSCIL evaluation metrics resemble those of CIL, primarily using overall classification accuracy (i.e., {\bf Overall acc.}) and, in some cases, Performance Dropping Rate (PD)\cite{DBLP:conf/cvpr/ZhangSLZPX21}. 
The PD is defined as $PD = \mathrm{Overall\ acc.}_0 - \mathrm{Overall\ acc.}_T$.

However, FSCIL, unlike CIL, features a higher proportion of base classes, leading accuracy in each session to be primarily influenced by these base classes. 
Consequently, PD becomes less effective for measuring forgetting as long as the base classes maintain high accuracy.
Recent methods\cite{DBLP:conf/cvpr/0001WYMPZ22,DBLP:conf/eccv/KallaB22,DBLP:journals/corr/abs-2312-05229} introduce harmonic mean between base classes and incremental classes,
but this measure remains affected by the high accuracy of base classes, i.e., the high accuracy of the base class still elevates the harmonic mean, lacking an intuitive reflection of the accuracy gap between overall and incremental classes.

Thus, we further define {\bf Base acc.}, {\bf Inc. acc.}, {\bf CInc. acc.}, and {\bf PInc. acc.} based on {\bf Overall acc.}, 
and define two accuracy ratios ({\bf Base/Inc.} and {\bf CInc./PInc.}) and {\bf BICP} in analyzing the accuracy imbalance phenomenon.
Specifically as follows:

{\bf Overall acc.} $\mathrm{Overall\ acc.}_{t}$ is the average classification accuracy of all seen classes, i.e., $\mathcal{C}^0\cup\mathcal{C}^1\cdots\cup\mathcal{C}^t$ in session $t$ ($t=0,...,T$), and  
\begin{equation}
  \mathrm{Overall\ acc.}_{\mathrm{Avg.}} = \frac{1}{T+1} \sum_{t=0}^{T} \mathrm{Overall\ acc.}_{t}.
  \label{eq:oa}
\end{equation}

{\bf Base acc.} $\mathrm{Base.\ acc.}_{t}$ is the average classification accuracy of all base classes $\mathcal{C}^0$ in session $t$ ($t=0,...,T$).

{\bf Inc. acc.} $\mathrm{Inc.\ acc.}_{t}$ is the average classification accuracy of all seen incremental classes, i.e., $\mathcal{C}^1\cdots\cup\mathcal{C}^t$ in session $t$  ($t=1,...,T$).

{\bf Base/Inc.} Accuracy ratio $\mathrm{Base/Inc.}$ is defined to measure the accuracy imbalance degree between base classes and incremental classes, and the closer the value is to 1, the closer it is to balance,
\begin{equation}
  \mathrm{Base/Inc.} = \frac{\mathrm{Base\ acc.}_{\mathrm{Avg.}}}{\mathrm{Inc.\ acc.}_{\mathrm{Avg.}}},
  \label{eq:oa}
\end{equation}
where
\begin{equation}
  \mathrm{Base\ acc.}_{\mathrm{Avg.}} = \frac{1}{T} \sum_{t=1}^{T} \mathrm{Base\ acc.}_{t}
  \label{eq:oa}
\end{equation}
and 
\begin{equation}
  \mathrm{Inc.\ acc.}_{\mathrm{Avg.}} = \frac{1}{T} \sum_{t=1}^{T} \mathrm{Inc.\ acc.}_{t}.
  \label{eq:oa}
\end{equation}

{\bf CInc. acc.} $\mathrm{CInc.\ acc.}_{t}$ is the average classification accuracy of all current incremental classes $\mathcal{C}^t$ in each session $t$ ($t=1,...,T$).

{\bf PInc. acc.} $\mathrm{PInc.\ acc.}_{t}$ is the average classification accuracy of all past incremental classes $\mathcal{C}^1\cdots\cup\mathcal{C}^{t-1}$ in each session $t$ ($t=2,...,T$). 

{\bf CInc./PInc.} Accuracy ratio $\mathrm{CInc./PInc.}$ is defined to measure the accuracy imbalance degree between current incremental classes and past incremental classes, and the closer the value is to 1, the closer it is to balance,
\begin{equation}
  \mathrm{CInc./PInc.} = \frac{\mathrm{CInc.\ acc.}_{\mathrm{Avg.}}}{\mathrm{PInc.\ acc.}_{\mathrm{Avg.}}},
  \label{eq:oa}
\end{equation}
where
\begin{equation}
  \mathrm{CInc.\ acc.}_{\mathrm{Avg.}} = \frac{1}{T-1} \sum_{t=2}^{T} \mathrm{CInc.\ acc.}_{t}
  \label{eq:oa}
\end{equation}
and 
\begin{equation}
  \mathrm{PInc.\ acc.}_{\mathrm{Avg.}} = \frac{1}{T-1} \sum_{t=2}^{T} \mathrm{PInc.\ acc.}_{t}.
  \label{eq:oa}
\end{equation}

{\bf BICP} To measure the degree of accuracy imbalance of the FSCIL method holistically, we calculated the mean for Base/Inc. and CInc./PInc. as BICP, and the closer the value is to 1, the more balanced the accuracy is.

\section{Additional Definitions}
\begin{definition}
\label{def:c}
(Class-specific Discriminative Features) 
Let $\phi_l$ be a feature mapping that transforms the input $\mathbf{x}_{i}$ into a feature $\phi_l(\mathbf{x}_{i})$, we say that $\phi_l(\mathbf{x}_{i})$ is a class-specific discriminative feature of classes $y_{i}\in \mathcal{C}^{b}$ if the following conditions hold:
For most $\mathbf{x}_{i}$ ($y_{i}\in \mathcal{C}^{b}$), $\phi'_l(\mathbf{x}_{i})$ is a distinct value close to 0 or 1, and there exists at least one $\mathbf{x}_{i}$ ($y_{i}\in \mathcal{C}^{b}$) such that $\phi'_l(\mathbf{x}_{i})$ approaches 1;
For most $\mathbf{x}_{i}$ ($y_{i}\notin \mathcal{C}^{b}$), $\phi'_l(\mathbf{x}_{i})$ is a random value between 0 and 1 in the chaotic state.

\end{definition}

\begin{definition}
\label{def:t}
(Transferable Features) 
Let $\phi_l$ be a feature mapping that transforms the input $\mathbf{x}_{i}$ into a feature $\phi_l(\mathbf{x}_{i})$, we say that $\phi_l(\mathbf{x}_{i})$ is a transferable feature of classes $y_{i}\in \mathcal{C}^{b}$ if the following conditions hold:
For any $\mathbf{x}_{i}$, $\phi'_l(\mathbf{x}_{i})$ is a distinct value between 0 and 1.
\end{definition}

\begin{definition}
\label{def:fmp}
(Feature Mapping Occupancy) 
Consider the definition of feature mappings in \cref{eq:g},
if feature $\phi_l(\mathbf{x}_{i})$ with a high activation value, the feature mapping $\phi_l$ is occupied to recognize the class $y_{i}$.
The occupancy degree of the feature mappings for recognizing sample $\mathbf{x}$ is defined by calculating the sum of all normalized features $\phi'_l(\mathbf{x})$, that is,
\begin{equation}
  \mathbf{FMO} = \sum_{l=1}^{d} \phi'_l(\mathbf{x}).
  \label{eq:act}
\end{equation}
\end{definition}
The larger the value of FMO, the more feature mappings (i.e., feature space) are occupied to recognized sample $\mathbf{x}$.

\section{Self-optimizing for Fine-grained Dataset}
\label{sec:BGMM}

To provide a more fine-grained description of classes, we further propose the improved classifiers $h_b$/$\tilde{h}_b$ based on Bayesian Gaussian mixture model (BGMM), where
$h_b=\{[p(\mathcal{X}|\theta_{c,1}),p(\mathcal{X}|\theta_{c,2})]| c\in {\textstyle \cup_{t=0}^{t'}} \mathcal{C}^{t}\}$.
Each data distribution (component) of class $c$ is fitted by BGMM as follows,
\begin{equation}
p(\mathcal{X}|\theta_{c,j})=\sum_{m=1}^M\pi_{c,j}^{m}\mathrm{N}\left(\mathcal{X}|\mu_{_{c,j}}^{m},\Sigma_{_{c,j}}^{m}\right), j=1,2,
\label{eq:bgmm}
\end{equation}
where $M$ is the number of distributions for the component $j$ of class $c$, $\mathrm{N}(\mathcal{X}|\mu_{_{c,j}}^{m},\Sigma_{_{c,j}}^{m})$ represents the probability density function of the $m$-th Gaussian distribution with means $\mu_{_{c,j}}^{m}$ and variances $\Sigma_{_{c,j}}^{m}$, 
and $\pi_{c,j}^{m}$ denotes the weight of the $m$-th Gaussian distribution, satisfying the condition $\sum_{m=1}^M\pi_{c,j}^{m}=1$ and $\pi_{c,j}^{m}\ge 0$.
The formula omits the prior term for simplicity.

In contrast to the generation of prototype based classifiers (i.e., calculating the mean vector of the training samples and their flipped counterpart belonging to a class respectively), 
BGMM based classifiers require prior specification of the maximum number $M$ of Gaussian distributions and the covariance matrix type (set to diag to reduce the storage cost). 
Then, the Expectation-Maximization (EM) algorithm \cite{1977Maximum} is employed to estimate the model parameters $\theta_{c,j}=\{\pi_{c,j}^{m},\mu_{_{c,j}}^{m},\Sigma_{_{c,j}}^{m}\}_{m=0}^{M}$ (encompassing means $\mu_{_{c,j}}^{m}$, covariances $\Sigma_{_{c,j}}^{m}$, and the weight $\pi_{c,j}^{m}$ for each Gaussian distribution), optimizing the likelihood of the training data. 

During the inference stage, overall mean vector for each BGMM needs to be computed,
\begin{equation}
\mu_{_{c,j}} = \sum_{m=1}^{M}\Sigma_{_{c,j}}^{m} \cdot \mu_{_{c,j}}^{m}, j=1,2.
\end{equation}
Subsequently, we replace the $P_c$ in \cref{eq:cls} with a set of mean vectors $[\mu_{_{c,1}}, \mu_{_{c,2}}]$ to achieve classification,
\begin{equation}
  y_{i}^\star=\underset{c\in {\textstyle \cup_{t=0}^{t'}} \mathcal{C}^{t}}{\operatorname{argmax}}\mathcal{S}(g(\mathbf{x}_i),[\mu_{_{c,1}}, \mu_{_{c,2}}]),
\end{equation}
where the test sample ${\mathbf{x}}_i$ adheres to the procedure employed in the prototype based classification, i.e., generating a set of feature vectors based on test sample ${\mathbf{x}}_i$ and its flipped counterpart $\hat{\mathbf{x}}_i$, as represented by the following formula: $g (\mathbf{x}_i)=[g (\mathbf{x}_i), g (\hat{\mathbf{x}}_i)]$.

{\em Resistance} is achieved within a certain range by continuously decaying the weights $\tilde{\pi}_{_{c,j}}^{k}$ of the $k$-th Gaussian distribution that is highly similar to the novel class $i$,
\begin{equation}
  \tilde{\pi}_{_{c,j}}^{k} \gets \gamma' \cdot (1-\mathcal{S} (\tilde{\mu}_{_{c,j}}^{k},\sum_{m'=1}^{M'}\tilde{\Sigma}_{_{i,j}}^{m'} \cdot \tilde{\mu}_{_{i,j}}^{m'})) \cdot \tilde{\pi}_{_{c,j}}^{k}, j=1,2,
 \label{eq:bca1}
\end{equation}
\begin{equation}
 k = \mathop{\arg \max}\limits_{m=1,...,M} \mathcal{S} (\tilde{\mu}_{_{c,j}}^{m},\sum_{m'=1}^{M'}\tilde{\Sigma}_{_{i,j}}^{m'} \cdot \tilde{\mu}_{_{i,j}}^{m'}), i \in \mathcal{C}^{t}, 
 \label{eq:bca2}
\end{equation}
where $\gamma'$ is a random number that controls the resistance degree. 
After weight decay, each weight $\tilde{\pi}_{_{c,j}}^{m}$ is divided by the sum of all weights so that their sum remains $1$.

{\em Calibration} is implemented by EM algorithm\cite{1977Maximum},
\begin{equation}
  \theta_{c,j} \gets EM(\theta_{c,j}, \mathcal{X}_{test}, \mu^{p}_{c,j}, \alpha' ), 
  \label{eq:ca3}
\end{equation}
where 
$\mu^{p}_{c,j}$ denotes the mean prior of the training set, $\alpha' $ is the regularization parameter for the mean.
To avoid distribution drift, the mean prior always uses $\mu^{p}_{c,j}$,
instead of calculating based on $\mathcal{X}_{test}$.

\section{Experimental Setup Details}
\label{sec:setup}
{\bf Datasets.}
We perform experiments on {\em mini}ImageNet, CIFAR100, and CUB200 datasets. 
{\em mini}ImageNet is a subset of the ImageNet dataset, comprising 600 images per class, with 500 allocated for training and 100 for testing purposes. Similarly, each class of CIFAR100 consists of 500 training images and 100 testing images. CUB200 is a fine-grained dataset comprising 200 classes, containing a total of 6000 training images and 6000 testing images.
The statistic characteristics of three datasets are listed in \cref{tab:data}.
\begin{table}[h]
  \centering
  \caption{ Statistics of datasets. $\left | \mathcal{C}^{0} \right | $: number of base classes. $T$: number of incremental sessions.
  }
  \begin{tabular}{lccccc}
    \toprule
    \multicolumn{1}{c}{Dataset}  & $\left | \mathcal{C}^{0} \right | $     & $T$     & $N$ & $K$ & Resolution \\
    \midrule
    {\em mini}ImageNet\cite{DBLP:journals/ijcv/RussakovskyDSKS15} & 60    & 8     & 5     & 5     &  84×84 \\
    CIFAR100\cite{Krizhevsky2009LearningML} & 60    & 8     & 5     & 5     & 32×32 \\
    CUB200\cite{Wah2011TheCB} & 100   & 10    & 10    & 5     & 224×224 \\
    \bottomrule
    \end{tabular}%
  \label{tab:data}
\end{table}

{\bf Implementation Details.} 
Our method is conducted with PyTorch library and SGD with momentum is used for optimization. 
The initial learning rate is set to 0.01 for {\em mini}ImageNet and CIFAR100 datasets, and 0.001 for CUB200 dataset. 
We adopt the standard data preprocessing including random resizing, random horizontal flipping, and color jittering in \cref{sec:f}. 
In \cref{eq:lamda}, $\lambda $ is constrained to a randomly selected value between $[0.4, 0.6]$ to minimize the overlap between virtual novel classes and real base classes.
Since CUB200 is a fine-grained dataset with subtle inter-class differences, inter-class fusion is not applied to it.
The output feature size of two fully connected layers of the SR module is 2048.
We set $r$ as 0.8 and $R$ as 40 in \cref{eq:ca2} for unlabeled sample recognition across all datasets. 
In BGMM based classifiers, $M$ in \cref{eq:bgmm} is set to $1$ for incremental classes and $3$ for base classes across all datasets.
Due to the typically lower bias associated with initial classifiers generated from larger sample sizes, the subsequent calibration degree should be lower. 
Specifically, the base class training samples used for classifier generation significantly outnumber the training samples available for the novel classes, and the number of base class training samples in the {\em mini}ImageNet and CIFAR100 datasets is notably higher than that in the CUB200 dataset. 
Additionally, datasets with larger intra-class variances often result in initial classifiers with higher biases, necessitating a higher degree of subsequent calibration.
The CUB200 dataset, as a fine-grained dataset, exhibits higher intra-class variance. 
In our method, $\alpha $ of prototype based classifiers in \cref{eq:ca1} and $\alpha' $ of BGMM based classifiers in \cref{eq:ca3} exert control over the degree of classifier calibration. 
As $\alpha $ increases (within the range $[0,1]$), the degree of classifier calibration becomes greater.
Larger value of $\alpha' $ (within the range $[0,+\infty]$) concentrates the cluster means around the mean prior $\mu^{p}_{c,j}$, i.e., a larger $\alpha' $ corresponds to a smaller degree of calibration.
Therefore, $\alpha $  is set to $0.1$ for base classes of {\em mini}ImageNet and CIFAR100, $0.6$ for base classes of CUB200 and incremental classes of {\em mini}ImageNet and CIFAR100,
and $0.9$ for incremental classes of CUB200. 
$\alpha' $ is set to $20$ for base classes of {\em mini}ImageNet and CIFAR100, $10$ for base classes of CUB200 and incremental classes of {\em mini}ImageNet and CIFAR100, and $2$ for incremental classes of CUB200. 

\section{Additional Experimental Results and Analyses}
\subsection{Additional comparison results}
\label{sec:results}
The detailed overall accuracy (Overall acc.) results for {\em mini}ImageNet, CIFAR100 and CUB200 datasets are shown in \cref{tab:mini,tab:cif,tab:cub}, which are omitted in \cref{tab:comp,fig:comp} of the main text. 
The BIPC for CIFAR100 and CUB200 are also listed.



\begin{table*}[h]
  \centering
  \caption{Comparison with SOTA methods on {\em mini}ImageNet dataset in terms of overall accuracy.
    }
    \begin{tabular}{lccccccccc}
    \toprule
    \multicolumn{1}{c}{\multirow{2}[4]{*}{Method}} & \multicolumn{9}{c}{Overall acc. (\%)} \\
\cmidrule{2-10}          & 0     & 1     & 2     & 3     & 4     & 5     & 6     & 7     & 8  \\
    \midrule
    TOPIC\cite{DBLP:conf/cvpr/TaoHCDWG20} & 61.31  & 50.09  & 45.17  & 41.16  & 37.48  & 35.52  & 32.19  & 29.46  & 24.42  \\
    CEC\cite{DBLP:conf/cvpr/ZhangSLZPX21}   & 72.00  & 66.83  & 62.97  & 59.43  & 56.70  & 53.73  & 51.19  & 49.24  & 47.63  \\
    FACT\cite{DBLP:conf/cvpr/0001WYMPZ22}  & 75.32  & 70.34  & 65.84  & 62.05  & 58.68  & 55.35  & 52.42  & 50.42  & 48.51  \\
    C-FSCIL\cite{DBLP:conf/cvpr/HerscheKCBSR22} & 76.40  & 71.14  & 66.46  & 63.29  & 60.42  & 57.46  & 54.78  & 53.11  & 51.41  \\
    TEEN\cite{DBLP:journals/corr/abs-2312-05229}  & 73.53  & 70.55  & 66.37  & 63.23  & 60.53  & 57.95  & 55.24  & 53.44  & 52.08  \\
    Bidist\cite{DBLP:conf/cvpr/Zhao0XC0NF23} & 74.65  & 70.43  & 66.29  & 62.77  & 60.75  & 57.24  & 54.79  & 53.65  & 52.22  \\
    FCIL\cite{DBLP:conf/ICCV/23}  & 76.34  & 71.40  & 67.10  & 64.08  & 61.30  & 58.51  & 55.72  & 54.08  & 52.76  \\
    SAVC\cite{DBLP:conf/cvpr/SongZSP0023}  & 81.12  & 76.14  & 72.43  & 68.92  & 66.48  & 62.95  & 59.92  & 58.39  & 57.11  \\
    NC-FSCIL\cite{DBLP:conf/iclr/YangYLLTT23} & 84.02  & 76.80  & 72.00  & 67.83  & 66.35  & 64.04  & 61.46  & 59.54  & 58.31  \\
    \midrule
    ours (Prototype) & \textbf{86.22} & \textbf{77.89} & \textbf{74.36} & \textbf{70.51} & \textbf{68.14} & \textbf{65.35} & \textbf{62.84} & 61.20  & 59.88  \\
    ours (BGMM) & \textbf{86.22}  & 77.38  & 73.90  & 70.13  & 67.85  & 65.11  & \textbf{62.84}  & \textbf{61.61} & \textbf{60.47} \\
    \bottomrule
    \end{tabular}%
  \label{tab:mini}%
\end{table*}%

\begin{table}[h]
  \centering
  \caption{Comparison with SOTA methods on CIFAR100 dataset in terms of overall accuracy and BIPC.}
    \begin{tabular}{lcccccccccc}
    \toprule
    \multicolumn{1}{c}{\multirow{2}[4]{*}{Method}} & \multicolumn{9}{c}{Overall acc. (\%)}                                 & \multirow{2}[4]{*}{BIPC} \\
\cmidrule{2-10}          & \textcolor[rgb]{ .2,  .2,  .2}{0} & \textcolor[rgb]{ .2,  .2,  .2}{1} & \textcolor[rgb]{ .2,  .2,  .2}{2} & \textcolor[rgb]{ .2,  .2,  .2}{3} & \textcolor[rgb]{ .2,  .2,  .2}{4} & \textcolor[rgb]{ .2,  .2,  .2}{5} & \textcolor[rgb]{ .2,  .2,  .2}{6} & \textcolor[rgb]{ .2,  .2,  .2}{7} & \textcolor[rgb]{ .2,  .2,  .2}{8} &  \\
    \midrule
    CEC   & 73.07  & 68.88  & 65.26  & 61.19  & 58.09  & 55.57  & 53.22  & 51.34  & 49.14  & 2.12  \\
    FACT  & 78.80  & 72.40  & 68.33  & 64.31  & 61.07  & 58.11  & 56.23  & 54.07  & 52.13  & 1.99  \\
    C-FSCIL & 77.47 & 72.4  & 67.47 & 63.25 & 59.84 & 56.95 & 54.42 & 52.47 & 50.47 & 2.99  \\
    TEEN  & 78.92  & 72.32  & 68.16  & 64.43  & 61.19  & 58.48  & 56.11  & 54.03  & 51.87  & 1.89  \\
    SAVC  & 78.47  & 72.31  & 67.49  & 62.41  & 59.10  & 55.95  & 53.81  & 51.54  & 49.16  & 1.97  \\
    NC-FSCIL & 82.52  & 76.82  & 73.34  & 69.68  & 66.19  & 62.85  & 60.96  & 59.02  & 56.11  & 1.66  \\
    \midrule
    Ours(Prototype) & \textbf{82.88} & \textbf{78.94} & \textbf{74.59} & \textbf{70.35} & \textbf{67.85} & \textbf{64.99} & \textbf{63.79} & \textbf{61.92} & \textbf{59.68} & \textbf{1.18} \\
    Ours(BGMM) & \textbf{82.88} & 78.77  & 73.89  & 69.73  & 67.21  & 64.42  & 63.44  & 61.57  & 59.40  & \textbf{1.18} \\
    \bottomrule
    \end{tabular}%
  \label{tab:cif}%
\end{table}%

\begin{table}[!h]
  \centering
  \caption{Comparison with SOTA methods on CUB200 dataset in terms of overall accuracy and BIPC.}
    \begin{tabular}{lcccccccccccc}
    \toprule
    \multicolumn{1}{c}{\multirow{2}[4]{*}{Method}} & \multicolumn{11}{c}{Overall acc. (\%)}                                                & \multirow{2}[4]{*}{BIPC} \\
\cmidrule{2-12}          & \textcolor[rgb]{ .2,  .2,  .2}{0} & \textcolor[rgb]{ .2,  .2,  .2}{1} & \textcolor[rgb]{ .2,  .2,  .2}{2} & \textcolor[rgb]{ .2,  .2,  .2}{3} & \textcolor[rgb]{ .2,  .2,  .2}{4} & \textcolor[rgb]{ .2,  .2,  .2}{5} & \textcolor[rgb]{ .2,  .2,  .2}{6} & \textcolor[rgb]{ .2,  .2,  .2}{7} & \textcolor[rgb]{ .2,  .2,  .2}{8} & 9     & 10    &  \\
    \midrule
    CEC   & 75.85  & 71.94  & 68.50  & 63.50  & 62.43  & 58.27  & 57.73  & 55.81  & 54.83  & 53.52  & 52.28  & 1.67  \\
    FACT  & 75.90  & 73.23  & 70.84  & 66.13  & 65.56  & 62.15  & 61.74  & 59.83  & 58.41  & 57.89  & 56.94  & 1.43  \\
    TEEN  & 77.26  & 76.13  & 72.81  & 68.16  & 67.77  & 64.40  & 63.25  & 62.29  & 61.19  & 60.32  & 59.31  & 1.32  \\
    SAVC  & \textbf{81.85} & 77.92  & 74.95  & 70.21  & 69.96  & 67.02  & 66.16  & 65.30  & 63.84  & 63.15  & 62.50  & 1.42  \\
    NC-FSCIL & 80.45  & 75.98  & 72.30  & 70.28  & 68.17  & 65.16  & 64.43  & 63.25  & 60.66  & 60.01  & 59.44  & 1.61  \\
    \midrule
    Ours(Prototype) & 81.22  & 78.05  & 75.28  & 71.49  & 70.18  & 67.45  & 66.30  & 65.53  & 64.10  & 63.83  & 63.15  & 1.28  \\
    Ours(BGMM) & 81.22  & \textbf{78.40} & \textbf{75.77} & \textbf{72.40} & \textbf{71.10} & \textbf{68.35} & \textbf{67.25} & \textbf{66.40} & \textbf{64.71} & \textbf{64.56} & \textbf{63.89} & \textbf{1.27} \\
    \bottomrule
    \end{tabular}%
  \label{tab:cub}%
\end{table}%

In addition, \cref{tab:mini1,tab:cif1,tab:cub1} showcase the incremental accuracy (Inc. acc.) in each session and Base/Inc. for our method and other recent SOTAs on {\em mini}ImageNet, CIFAR100 and CUB200 datasets.

\begin{table}[h]
  \centering
  \caption{Comparison with SOTA methods on {\em mini}ImageNet dataset in terms of incremental accuracy and Base/Inc..}
    \begin{tabular}{lccccccccc}
    \toprule
    \multicolumn{1}{c}{\multirow{2}[4]{*}{Method}} & \multicolumn{8}{c}{Inc. acc. (\%)}                            & \multirow{2}[4]{*}{Base/Inc.} \\
\cmidrule{2-9}          & \textcolor[rgb]{ .2,  .2,  .2}{1} & \textcolor[rgb]{ .2,  .2,  .2}{2} & \textcolor[rgb]{ .2,  .2,  .2}{3} & \textcolor[rgb]{ .2,  .2,  .2}{4} & \textcolor[rgb]{ .2,  .2,  .2}{5} & \textcolor[rgb]{ .2,  .2,  .2}{6} & \textcolor[rgb]{ .2,  .2,  .2}{7} & \textcolor[rgb]{ .2,  .2,  .2}{8} &  \\
    \midrule
    CEC   & 15.20  & 17.00  & 16.67  & 16.05  & 14.32  & 13.73  & 14.31  & 14.88  & 4.59  \\
    FACT  & 15.80  & 14.40  & 15.40  & 14.55  & 13.64  & 12.20  & 12.66  & 13.20  & 5.30  \\
    C-FSCIL & 5.20  & 12.40  & 17.27  & 19.90  & 23.40  & 22.60  & 25.91  & 25.95  & 3.83  \\
    TEEN  & 40.20  & 35.60  & 32.47  & 32.70  & 29.96  & 28.33  & 28.89  & 29.35  & 2.17  \\
    Bidist & 27.00  & 30.30  & 29.60  & 27.85  & 28.44  & 26.80  & 26.23  & 25.62  & 2.51  \\
    SAVC  & 33.80  & 30.30  & 29.67  & 30.50  & 27.32  & 25.20  & 25.46  & 26.58  & 2.72  \\
    NC-FSCIL & 54.20  & 48.10  & 42.73  & 40.55  & 34.88  & 32.00  & 32.09  & 31.33  & 1.92  \\
    \midrule
    Ours(Prototype) & \textbf{60.00} & \textbf{54.20} & \textbf{48.33} & \textbf{48.05} & \textbf{44.64} & \textbf{43.40} & \textbf{44.51} & \textbf{45.35} & \textbf{1.53} \\
    Ours(BGMM) & 59.40  & 54.00  & \textbf{48.33} & 47.95  & 44.16  & 42.57  & 43.23  & 44.23  & 1.55  \\
    \bottomrule
    \end{tabular}%
  \label{tab:mini1}%
\end{table}%

\begin{table}[h]
  \centering
  \caption{Comparison with SOTA methods on CIFAR100 dataset in terms of incremental accuracy and Base/Inc..}
    \begin{tabular}{lccccccccc}
    \toprule
    \multicolumn{1}{c}{\multirow{2}[4]{*}{Method}} & \multicolumn{8}{c}{Inc. acc. (\%)}                            & \multirow{2}[4]{*}{Base/Inc.} \\
\cmidrule{2-9}          & \textcolor[rgb]{ .2,  .2,  .2}{1} & \textcolor[rgb]{ .2,  .2,  .2}{2} & \textcolor[rgb]{ .2,  .2,  .2}{3} & \textcolor[rgb]{ .2,  .2,  .2}{4} & \textcolor[rgb]{ .2,  .2,  .2}{5} & \textcolor[rgb]{ .2,  .2,  .2}{6} & \textcolor[rgb]{ .2,  .2,  .2}{7} & \textcolor[rgb]{ .2,  .2,  .2}{8} &  \\
    \midrule
    CEC   & 27.40  & 24.50  & 21.13  & 19.95  & 20.48  & 20.60  & 20.09  & 19.35  & 3.24  \\
    FACT  & 31.20  & 28.40  & 24.33  & 23.00  & 22.44  & 23.43  & 22.29  & 21.55  & 2.97  \\
    C-FSCIL & 18.00  & 13.60  & 13.00  & 12.65  & 15.36  & 16.30  & 15.74  & 16.73  & 4.92  \\
    TEEN  & 33.00  & 30.10  & 27.33  & 26.00  & 25.56  & 25.00  & 24.20  & 23.60  & 2.81  \\
    SAVC  & 35.60  & 28.50  & 25.93  & 25.75  & 26.96  & 27.53  & 26.57  & 26.65  & 2.49  \\
    NC-FSCIL & 44.00  & 41.60  & 36.47  & 31.95  & 31.32  & 33.97  & 31.31  & 29.30  & 2.19  \\
    \midrule
    Ours(Prototype) & \textbf{63.80} & 61.90  & 51.47  & 50.60  & 49.72  & 51.13  & 48.80  & \textbf{47.15} & 1.38  \\
    Ours(BGMM) & 63.40  & \textbf{62.60} & \textbf{51.93} & \textbf{50.80} & \textbf{49.88} & \textbf{51.40} & \textbf{48.91} & 47.00  & \textbf{1.36} \\
    \bottomrule
    \end{tabular}%
  \label{tab:cif1}%
\end{table}%

\begin{table}[!h]
  \centering
  \caption{Comparison with SOTA methods on CUB200 dataset in terms of incremental accuracy and Base/Inc..}
    \begin{tabular}{lccccccccccc}
    \toprule
    \multicolumn{1}{c}{\multirow{2}[3]{*}{Method}} & \multicolumn{10}{c}{Inc. acc. (\%)}                                           & \multirow{2}[3]{*}{Base/Inc.} \\
\cmidrule{2-11}          & \textcolor[rgb]{ .2,  .2,  .2}{1} & \textcolor[rgb]{ .2,  .2,  .2}{2} & \textcolor[rgb]{ .2,  .2,  .2}{3} & \textcolor[rgb]{ .2,  .2,  .2}{4} & \textcolor[rgb]{ .2,  .2,  .2}{5} & \textcolor[rgb]{ .2,  .2,  .2}{6} & \textcolor[rgb]{ .2,  .2,  .2}{7} & \textcolor[rgb]{ .2,  .2,  .2}{8} & \textcolor[rgb]{ .2,  .2,  .2}{9} & \textcolor[rgb]{ .2,  .2,  .2}{10} &  \\
   \midrule
    CEC   & 40.86  & 37.46  & 30.91  & 33.73  & 31.11  & 32.59  & 32.77  & 31.34  & 33.11  & 32.95  & 2.18  \\
    FACT  & 52.33  & 47.02  & 37.61  & 39.70  & 37.59  & 39.25  & 39.79  & 38.25  & 40.14  & 39.74  & 1.80  \\
    TEEN  & 57.71  & 52.56  & 45.55  & 47.41  & 44.97  & 46.10  & 45.62  & 43.54  & 44.94  & 44.84  & 1.59  \\
    SAVC  & 51.61  & 50.15  & 42.81  & 45.76  & 43.38  & 45.18  & 46.41  & 44.85  & 46.58  & 47.35  & 1.69  \\
    NC-FSCIL & \textbf{66.67} & 45.41  & 42.59  & 45.88  & 41.72  & 44.11  & 44.99  & 41.16  & 42.66  & 43.07  & 1.68  \\
    \midrule
    Ours(Prototype) & 65.23  & 57.83  & 48.27  & 50.94  & 48.05  & 49.65  & 50.86  & 49.76  & 51.11  & 51.48  & 1.48  \\
    Ours(BGMM) & 66.31  & \textbf{58.75} & \textbf{50.60} & \textbf{53.07} & \textbf{49.93} & \textbf{51.16} & \textbf{52.16} & \textbf{50.46} & \textbf{51.89} & \textbf{52.05} & \textbf{1.45} \\
    \bottomrule
    \end{tabular}%
  \label{tab:cub1}%
\end{table}%

\subsection{The analyses of \cref{fig:vis1}.}
\label{sec:vis1}
To make the results clearer, we show feature vectors $g(\mathbf{x})$ of all base classes and randomly selected five incremental classes with t-SNE in \cref{fig:vis1}.
It can be observed that incremental class samples in the baseline are mostly mapped to the base class positions (see \cref{fig:visa}). 
The intra-class transformation and inter-class fusion make it possible for incremental class samples to be mapped to the feature space positions that are not occupied by base classes, resulting in a clustering trend for incremental classes (see \cref{fig:visb} and \cref{fig:visc}).
However, incremental class samples cannot cluster well due to the lack of effective discriminative features. 
Adding the SR module after $g$ further makes incremental classes form effective clusters in the feature space, but it blurs the boundaries among base classes and between base classes and incremental classes (see \cref{fig:visd}), 
so we propose the separately dual-feature classification to intelligently 
combine $g (\mathbf{x})$ with $\tilde{g} (\mathbf{x})$.

\subsection{The results of FMO.}
\label{sec:fmo}
\begin{figure}[!h]
  \centering
   \includegraphics[width=0.6\linewidth]{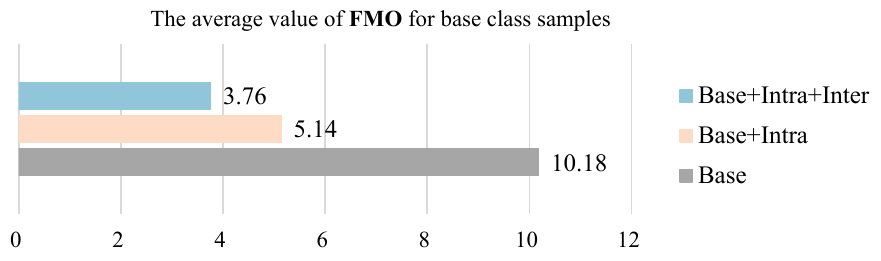}
   \caption{Comparison of the average value of {\bf FMO} for base class samples on CIFAR100 test set. 
   }
   \label{fig:L1}
\end{figure}

We compare the changes of the average value of FMO for base class samples on CIFAR100 test set after introducing intra-class transformation (Intra) and inter-class fusion (Inter) in \cref{fig:L1}. 

\subsection{The analyses of \cref{fig:heatmap}.}
\label{sec:fig7}
Specifically, more features are activated in \cref{fig:hb}, indicating that adding $SR$ after $g$ indeed stimulates $g$ to learn and retain more features. 
Moreover, there is no obvious difference in the activation density and intensity of pixels between the first 6000 base class samples and the last 4000 incremental class samples.
According to \cref{def:t}, these features are transferable features, suggesting that these features are basically not biased towards base classes, i.e., 
the number and ability of feature mappings used for learning base and incremental classes are similar.

In \cref{fig:ha}, 
the activation density of pixels for base classes is notably lower, with generally higher or lower activation values.
This implies that the number of feature mappings occupied by the base class is compressed through intra-class transformation and inter-class fusion, 
and the few feature mappings utilized for the base classes exhibit a superiority in discriminative ability compared to other feature mappings.
This is because other feature mappings deemed to interfere with the recognition of base classes, yet applicable for recognizing incremental classes, have been weakened or abandoned by the feature extractor throughout the base training process, i.e., the mapping results of these feature mappings approach zero for all classes. 
Consequently, we preserve transferable features before $SR$ to complement the final class-specific discriminative features.

\begin{table*}[ht]
  \centering
  \caption{Overall accuracy and incremental accuracy of our separately dual-feature classification compared to other ways to utilizing transferable features $g(\mathbf{x})$ and class-specific discriminative features $\tilde{g}(\mathbf{x})$ on CIFAR100 dataset.
    }
    \begin{tabular}{l|ccccccccc}
    \toprule
    \multicolumn{1}{c|}{\multirow{2}[4]{*}{Method}} & 0     & 1     & 2     & 3     & 4     & 5     & 6     & 7     & 8 \\
\cmidrule{2-10}          & \multicolumn{9}{c}{Overall acc. (\%)} \\
    \midrule
    $g(\mathbf{x})$  & 79.57  & 76.20  & 72.99  & 69.01  & 66.93  & 64.12  & 62.90  & 61.15  & 58.89  \\
    $\tilde{g}(\mathbf{x})$ & 82.88  & 77.82  & 72.57  & 68.24  & 64.61  & 61.78  & 58.78  & 56.18  & 53.75  \\
    \textbf{Pre} & \textbf{82.92}  & 77.91  & 72.63  & 68.32  & 64.76  & 61.89  & 59.03  & 56.42  & 54.07  \\
    \textbf{Post} & 78.82  & 74.38  & 70.19  & 67.03  & 63.99  & 60.91  & 58.63  & 56.43  & 54.77  \\
    \textbf{AD} & 82.88  & 78.46  & 73.64  & 69.36  & 66.35  & 63.49  & 61.39  & 59.34  & 57.02  \\
    \midrule
    Ours  & 82.88  & \textbf{78.89}  & \textbf{74.90}  & \textbf{70.63}  & \textbf{68.26}  & \textbf{65.19}  & \textbf{63.67}  & \textbf{61.84}  & \textbf{59.43}  \\
    \midrule
          & \multicolumn{9}{c}{Inc. acc. (\%)} \\
    \midrule
    $g(\mathbf{x})$  & -    & 41.00  & 38.30  & 31.87  & 33.55  & 32.48  & 35.30  & 34.94  & 33.00  \\
    $\tilde{g}(\mathbf{x})$ & -   & 30.80  & 22.10  & 19.93  & 18.20  & 19.44  & 18.30  & 17.20  & 17.22  \\
    \textbf{Pre} & -   & 31.20  & 22.40  & 20.27  & 18.80  & 19.96  & 19.00  & 17.80  & 17.93  \\
    \textbf{Post} & -   & 21.20  & 18.40  & 19.87  & 19.50  & 17.92  & 18.27  & 18.06  & 18.70  \\
    \textbf{AD} & -  & 24.60  & 18.60  & 16.80  & 17.80  & 18.72  & 20.17  & 20.57  & 19.77  \\
    \midrule
    Ours  & -   & \textbf{50.00}  & \textbf{44.70}  & \textbf{37.47}  & \textbf{38.05}  & \textbf{36.96}  & \textbf{39.13}  & \textbf{38.20}  & \textbf{36.65}  \\
    \bottomrule
    \end{tabular}%
    
  \label{tab:o2f}
\end{table*}

\subsection{The detailed results and analyses of \cref{tab:2f1}.}
\label{sec:tab3}
The detailed results are represented in \cref{tab:o2f}, encompassing the overall accuracy (Overall acc.) and incremental accuracy (Inc. acc.) in each session for our separately dual-feature classification strategy and other alternative ways that utilize transferable features $g(\mathbf{x})$ and class-specific discriminative features $\tilde{g}(\mathbf{x})$.
Among them, 
$\tilde{g}(\mathbf{x})$ performs well in the base session (base classes), whereas $g(\mathbf{x})$ exhibits superior performance in the incremental sessions (incremental classes).
No matter whether it is pre-integration (\textbf{Pre}, i.e., feature vector integration), post-integration (\textbf{Post}, i.e., similarity integration), or the idea of anomaly detection (\textbf{AD}, i.e., first use class-specific discriminative features to detect samples that do not belong to base classes, and then use their transferable features to reclassified them), they all get a low incremental accuracy similar to using class-specific discriminative features $\tilde{g}(\mathbf{x})$ alone. 
Our separately dual-feature classification idea can not only achieve the highest overall accuracy but also the highest incremental accuracy.


\end{document}